\documentclass[10pt,twocolumn,letterpaper]{article}

\usepackage{cvpr}
\usepackage{times}
\usepackage{epsfig}
\usepackage{graphicx}
\usepackage{amsmath}
\usepackage{amssymb}

\usepackage{algorithm}
\usepackage{algorithmic}
\usepackage{varwidth}
\usepackage{makecell}
\usepackage{graphicx}
\usepackage{subfig}
\usepackage{comment}
\usepackage{booktabs}
\usepackage{multirow}
\usepackage{graphicx}
\usepackage[toc,page]{appendix}
\newcolumntype{K}[1]{>{\centering\arraybackslash}p{#1}}
\let\emptyset\varnothing
\usepackage[breaklinks=true,bookmarks=false]{hyperref}

\usepackage[usenames,dvipsnames,svgnames,table]{xcolor}

\makeatletter
\def\mathcolor#1#{\@mathcolor{#1}}
\def\@mathcolor#1#2#3{%
  \protect\leavevmode
  \begingroup
    \color#1{#2}#3%
  \endgroup
}
\makeatother

\cvprfinalcopy 

\def\cvprPaperID{2548} 

\ifcvprfinal\fi
\begin{document}


\title{Creativity Inspired Zero-Shot Learning}


\author{Mohamed Elhoseiny$^{1,2}\,\,\,\,\,\,\,\,\,\,\,$ Mohamed Elfeki$^{3}$ \\
$^1${Facebook AI Research (FAIR)}, $^2${King Abdullah University of Science and Technology (KAUST)}\\
$^3${University of Central Florida}\\
{\tt\small mohamed.elhoseiny@kaust.edu.sa,elfeki@cs.ucf.edu}
}

\maketitle

\begin{abstract}


Zero-shot learning (ZSL) aims at understanding unseen categories with no training examples from class-level descriptions. To improve the discriminative power of zero-shot learning,  
we model the visual learning process of unseen categories with an inspiration from the psychology of human creativity for producing novel art. 
We relate ZSL to human creativity by observing that 
zero-shot learning is about recognizing the unseen and creativity is about creating a likable unseen. We introduce a learning signal inspired by creativity literature that explores the unseen space with hallucinated class-descriptions and encourages careful deviation of their visual feature generations from seen classes while allowing knowledge transfer from seen to unseen classes. 
Empirically, we show consistent improvement over the state of the art of several percents on the largest available benchmarks on the challenging task or generalized ZSL from a noisy text that we focus on, using the CUB and NABirds datasets. We also show the advantage of our approach on Attribute-based ZSL on three additional datasets (AwA2, aPY, and SUN). Code is available at  \url{https://github.com/mhelhoseiny/CIZSL}. 
\end{abstract}

\vspace{-2mm}
\section{Introduction}
With hundreds of thousands of object categories in the real world and countless undiscovered species, it becomes unfeasible to maintain hundreds of examples per class to fuel the training needs of most existing recognition systems. 
Zipf's law, named after George Zipf (1902$-$1950), suggests that for the vast majority of the world-scale classes, only a few examples are available for training, validated earlier in language (e.g., ~\cite{zipf1935human,zipf1949human}) and later in vision (e.g., ~\cite{salakhutdinov2011learning}).  This problem becomes even more severe when we target recognition at the fine-grained level. For example, there exists tens of thousands of bird and flower species, but the largest available benchmarks have only a few hundred classes motivating  a lot of research on classifying instances of unseen classes,  known as Zero-Shot Learning (ZSL). 

\begin{figure}[t!]
	\centering
	\vspace{-3mm}
	\includegraphics[width=8.3cm]{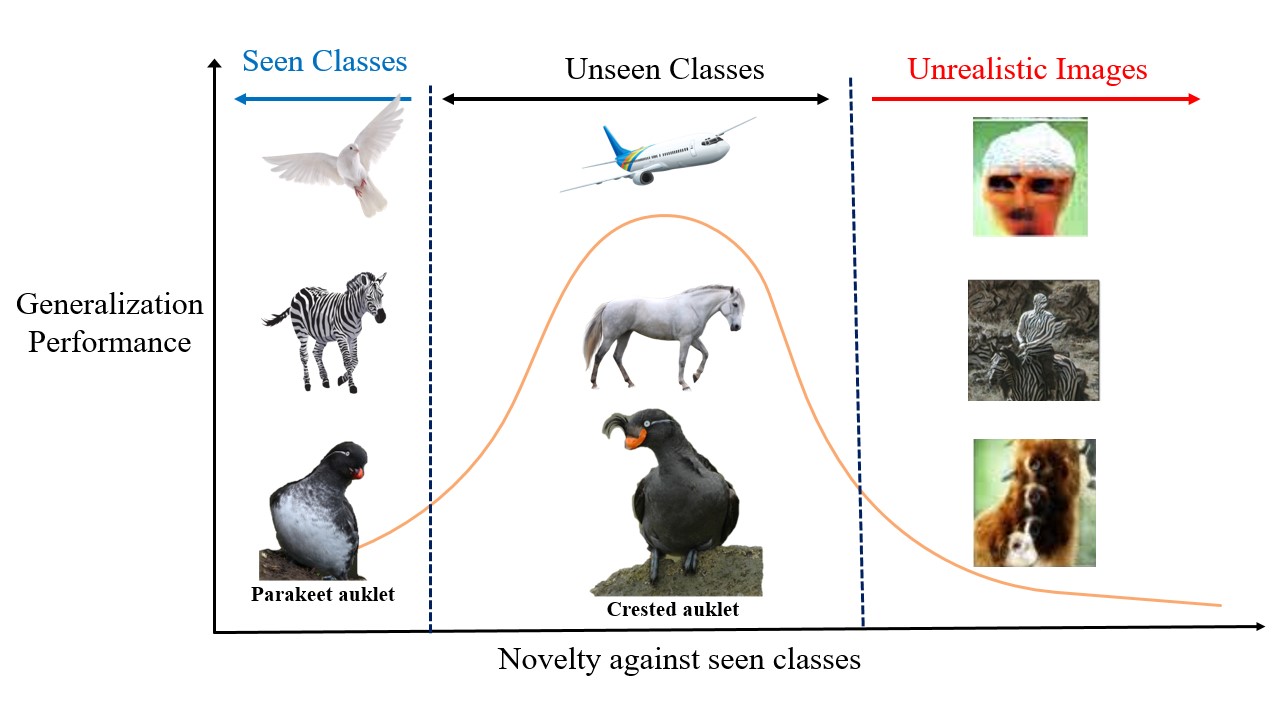}
		\vspace{-9mm}
	\caption{
	Generalizing the learning of zero-shot models requires a deviation from seen classes to accommodate recognizing unseen classes. 
	We carefully model a learning signal that inductively encourages deviation of unseen classes  from seen classes, yet not pushed far that the generation fall in the negative hedonic unrealistic range on the right and loses knowledge transfer from seen classes. Interestingly, this curve is similar to the famous Wundt Curve in the Human Creativity literature (Martindale, 1990) ~\cite{martindale1990clockwork}.}
	\label{fig:intro}
	\vspace{-6.5mm}
\end{figure}

People have a great capability to identify unseen visual classes from text descriptions like
{``The crested auklet is subspecies of birds with dark-gray bodies tails and wings and orange-yellow bill. It is known for its forehead crests, made of black forward-curving feathers.''}; see Fig~\ref{fig:intro} (bottom). We may \emph{imagine} the appearance of ``crested auklet'' in different ways yet all are correct and may collectively help us understand it better.
This  \emph{imagination} notion been modeled in recent ZSL approaches (e.g.,~\cite{guo2017synthesizing, long2017zero, guo2017zero,Elhoseiny_2018_CVPR,xian2018feature}) successfully adopting deep generative models to synthesize visual examples of an unseen object given its semantic description. After training, the model generates imaginary data for each unseen class transforming ZSL into a standard classification task with the generated data. 

However, these generative ZSL methods do not guarantee the discrimination between seen and unseen classes since the generations are not motivated with a learning signal to deviate from seen classes. For example, ``Parakeet Auklet''  as a seen class in Fig~\ref{fig:intro} (left) has a visual text description~\cite{wiki_crested19} that significantly overlaps with ``Crested Auklet'' description, yet one can identify ``Crested Auklet'''s unique ``'black forward-curving feathers'' against ``Parakeet Auklet'' from text. 
\emph{The core of our work is to address the question of how to produce discriminative generations of unseen visual classes from text descriptions  by explicitly learning to deviate from seen classes while allowing transfer to unseen classes.}  
Let's imagine the space of conditional visual generations from class descriptions on an intensity map where light regions implies seen and darker regions implies unseen.  These class descriptions are represented in a shared space between the unseen (dark) and the seen (light) classes, and hence the transfer is expected.  In existing methods, this transfer  signal is formulated by encouraging the generator to produce quality examples  conditioned only on  the descriptions of the seen classes ( light regions only). In this inductive zero-shot learning, class descriptions of unseen classes  are not available during training and hence can not used as a learning signal to explicitly encourage the discrimination across unseen and seen classes. \emph{Explicitly modeling an inductive and discriminative learning signal from the dark unseen space is at the heart of our work.}

\noindent \textbf{Creativity Inspiration to Zero-shot Learning.} We propose to extend generative zero-shot learning with a discriminative learning signal inspired from the psychology of human creativity. 
Colin Marindale~\cite{martindale1990clockwork}  proposes a psychological theory to explain the perception of human creativity. The definition relates likability of an art piece to novelty by {\it ``the principle of least effort''}. The aesthetic appeal of an art work first increases when it deviates from existing work till some point, then  decreases when the deviation goes too far. This means that it gets difficult to connect this art to what we are familiar with, and hence deems it hard to understand and hence appreciate. This principle can be visualized by the Wundt Curve where the X axis represents novelty and Y axis represents likability like an inverted U-shape; similar to the curve in Fig~\ref{fig:intro}. 
We relate the Wundt curve behavior in producing creative art to a desirable generalized ZSL mode that has a better capability to distinguish the ``crested auklet'' unseen class from the ``parakeet auklet'' seen class given how similar they are as mentioned before; see Fig~\ref{fig:intro}. A generative ZSL model that cannot deviate generations of unseen classes from instances of seen classes is expected to underperform in generalized zero-shot recognition due to confusion; see Fig~\ref{fig:intro}(left). As the deviation capability increases, the performance is expected to get better but similarly would decrease when the deviation goes too far producing unrealistic generation and reducing the needed knowledge transfer from seen classes; see Fig~\ref{fig:intro}(middle and right). Our key question is how to properly formulate deviation from generating features similar to existing classes while balancing the desirable transfer learning signal.

\noindent \textbf{Contributions.} 
{\noindent \textbf{1)} We propose a  zero-shot learning approach that explicitly models generating unseen classes by learning to carefully deviate from seen classes. We examine a parametrized entropy measure to facilitate learning how to deviate from seen classes. Our approach is inspired from the psychology of human creativity; and thus we name it Creativity Inspired Zero-shot Learning (CIZSL). 

\noindent \textbf{2)} Our creativity inspired loss is unsupervised and orthogonal to any Generative ZSL approach. Thus it can be integrated with any GZSL while adding no extra parameters nor requiring any additional labels. 

\noindent \textbf{3)} By means of extensive experiments on seven benchmarks encompassing Wikipedia-based and attribute-based descriptions, our approach consistently outperformed state-of-the-art methods on zero-shot recognition, zero-shot retrieval, and generalized zero-shot learning using several evaluation metrics.}

\vspace{-2mm}
\section{Related Work}
\noindent\textbf{Early Zero-Shot Learning(ZSL) Approaches} 
 A key idea to facilitate zero-shot learning is finding a common semantic representation that both seen and unseen classes can share. Attributes and text descriptions are shown to be effective shared semantic representations that allow transferring knowledge from seen to unseen classes.
 Lampert \emph{et al.}~\cite{lampert2009} proposed a Direct Attribute Prediction (DAP) model that assumed independence of attributes and estimated the posterior of the test class by combining the attribute prediction probabilities. A parallelly developed, yet similar model was developed by Farhadi \emph{et al.}~\cite{farhadi2009describing}.
 
 
 
\noindent \textbf{Visual-Semantic Embedding ZSL. }Relaxing the unrealistic independence assumption, Akata \emph{et al.}~\cite{akata2016label} proposed an Attribute Label Embedding(ALE) approach that models zero-shot learning as a linear joint visual-semantic embedding. In principal, this model is similar to prior existing approaches that learn a mapping function from visual space to semantic space~\cite{zhang2016learning, shigeto2015ridge}. This has been also investigated in the opposite direction~\cite{zhang2016learning, shigeto2015ridge} as well as jointly learning a function for each space that map to a common space~\cite{yang2014unified, lei2015predicting, akata2015evaluation, romera2015embarrassingly,socher2013zero,Elhoseiny_2017_CVPR,akata2016multi,long2017learning,long2017describing,tsai2017learning}.


\noindent\textbf{Generative ZSL Approaches}
The notion of generating artificial examples has been recently proposed to model zero-shot learning reducing it to a conventional classification problem~\cite{guo2017synthesizing, long2017zero, guo2017zero,Elhoseiny_2018_CVPR}. Earlier approaches assumed a Gaussian distribution prior for visual space to every class and the probability densities for unseen classes are modeled as a linear combination of seen class distributions~\cite{guo2017synthesizing}.  Long \emph{et al.}~\cite{long2017zero} instead proposed a one-to-one mapping approach where synthesized examples are restricted.  Recently, Zhu~\emph{et al.}~\cite{Elhoseiny_2018_CVPR},  Xian~\emph{et al.}~\cite{xian2018feature}, and Verma~\emph{et al.}\cite{kumar2018generalized} relaxed this assumption and built on top of  generative adversarial networks (GANs)~\cite{goodfellow2014generative,radford2015unsupervised} to generate examples from unseen class descriptions. Different from ACGAN~\cite{odena2016conditional}, Zhu \emph{et al.} added a visual pivot regularizer (VPG) encourages generations of each class to be close to the average of its corresponding real features.


\noindent\textbf{Semantic Representations in ZSL (e.g., Attributes, Description).}
ZSL requires by definition additional information (e.g., semantic description of unseen classes) to enable their recognition. A considerable progress has been made in studying attribute representation~\cite{lampert2009, Lampert2014, akata2016label,frome2013devise, zhang2016learning, yang2014unified, lei2015predicting, akata2015evaluation, romera2015embarrassingly, akata2016multi}. Attributes are a collection of semantic characteristics that are filled to uniquely describe unseen classes. Another ZSL trend is to use online textual descriptions ~\cite{elhoseiny2013write,Elhoseiny_2017_CVPR, Qiao2016, reed2016learning, lei2015predicting}. 
Textual descriptions  can be easily extracted from online sources like Wikipedia with a minimal overhead, avoiding the need to define hundreds of attributes and filling them for each class/image. 
Elhoseiny~\emph{et al.}~\cite{elhoseiny2013write} proposed an early approach for Wikipedia-based zero-shot learning that combines domain transfer and regression to predict visual classifiers from a TF-IDF textual representation~\cite{salton1988term}. Qiao~\emph{et al.}~\cite{Qiao2016} proposed  suppress the noise in the Wikipedia articles by encouraging sparsity of the neural weights to the text terms. Recently, part-based zero-shot learning model~\cite{Elhoseiny_2017_CVPR} was proposed  with a capability to connect text terms to its relevant parts of objects without part-text annotations. 
More recently, Zhu~\emph{et al.}~\cite{Elhoseiny_2018_CVPR} showed that suppressing the non-visual information is possible by the predictive power of the their model to synthesize visual features from the noisy Wikipedia text. 
Our work also focus on the challenging task of recognizing objects based on Wikipedia articles and is also a generative model. Unlike existing, we explicitly model the careful deviation of unseen class generations from seen classes .







\noindent \textbf{Visual Creativity.} 
Computational Creativity studies building machines that generate original items with realistic and aesthetic characteristics~\cite{machado2000nevar,mordvintsev2015inceptionism,dipaola2009incorporating}. Although GANs~\cite{goodfellow2014generative,radford2015unsupervised,ha2017neural} are a powerful generative model, yet it is not explicitly trained to create novel content beyond the training data. For instance, a GAN model trained on art works  might generate the ``Mona Lisa'' again, but would not produce a novel content that it did not see. It is not different for some existing style transfer work~\cite{Gatys2016ImageStyleTransfer,dumoulin2016learned} since there is no incentive in these models to generate a new content. More recent work 
adopts computational creativity literature to create novel art and fashion designs~\cite{elgammal2017can,sbai2018design}. Inspired by~\cite{martindale1990clockwork},   Elgammal\etal~\cite{elgammal2017can} adapted GANs to generate unconditional creative content (paintings) by encouraging the model to deviate from existing painting styles. 
Fashion is a 2.5 trillion dollar industry and has an impact in our everyday life, this motivated~\cite{sbai2018design} to develop a model that can for example create an unseen fashion shape ``pants to extended arm sleeves''. 
The key idea behind these models is to add an additional novelty loss that encourage the model to explore the creative space of image generation.

\vspace{-2mm}
\section{Background}
\label{sec_bg}
\vspace{-2mm}
GANs~\cite{goodfellow2014generative,radford2015unsupervised} train the generator $G$, with parameters $\theta_G$, to produce samples that the Discriminator $D$ believe they are real. On the other hand, the Discriminator $D$, with parameters $\theta_D$, is trained to classify samples from the real distribution $p_{data}$ as real (1), and samples produced by the generator as fake (0); see Eq~\ref{GAN_org}.
\begin{equation}
\vspace{-5mm}
\small
  \min_{\theta_G} \mathcal{L}_{G} = \min_{\theta_G} \sum_{z_i\in \mathbb{R}^n} \log (1-D(G(z_i))) \\
\end{equation}
\vspace{1em}
\begin{equation}
\small
  \min_{\theta_D} \mathcal{L}_{D} =\min_{\theta_D} \sum_{\substack{x_i}\in\mathcal{D}, z_i\in \mathbb{R}^n} - \log D(x_i) - \log (1-D(G(z_i))))
    \label{GAN_org}
\end{equation}
where $z_i$ is a noise vector sampled from prior distribution $p_{z}$ and $x$ is a real sample from the data distribution $p_{data}$.
In order to learn to deviate from seen painting styles or fashion shapes, \cite{elgammal2017can,sbai2018design} proposed an additional head for the discriminator $D$ that predicts the class of an image (painting style or shape class). During training, the Discriminator $D$ is trained to predict the class of the real data through its additional head, apart from the original real/fake loss.  The generator $G$ is then trained to generate examples that are not only classified as real but more importantly are encouraged to be hard to classify using the additional discriminator head. More concretely, 
\begin{equation}
\small 
  \mathcal{L}_{G} = \mathcal{L}_{G\mbox{ \scriptsize{real/fake}}} + \lambda  \mathcal{L}_{G\mbox{ \scriptsize{creativity}}} 
\end{equation}
The common objective between~\cite{elgammal2017can} and ~\cite{sbai2018design} is to produce novel generations with high entropy distribution over existing classes but they are different in the loss function. 
In~\cite{elgammal2017can}, $\mathcal{L}_{G\mbox{ \scriptsize{creativity}}}$ is defined as the binary cross entropy (BCE) over each painting style produced by the discriminator additional head and the uniform distribution (i.e., $\frac{1}{K}$, $K$ is the number of classes). Hence, this loss is a summation of BCE losses over all the classes. In contrast, Sbai~\emph{et al.}~\cite{sbai2018design} adopted the Multiclass Cross Entropy (MCE) between the distribution over existing classes and the uniform distribution. To our knowledge, creative generation has not been explored before conditioned on text and to also facilitate recognizing unseen classe, \emph{two key differences to our work.} 
{
Relating computational creativity to zero-shot learning is one of the novel aspects in our work by encouraging the deviation of generative models from seen classes. However, proper design of the learning signal is critical to (1) hallucinate class text-descriptions whose visual generations can help the careful deviation, (2) allow discriminative generation while allowing transfer between seen and unseen classes to facilitate zero-shot learning.
}

\vspace{-2mm}
\section{Proposed Approach}
\begin{figure*}[t!]
\vspace{-3mm}
	\centering
	\includegraphics[width=14.0cm,  height=5.5cm]{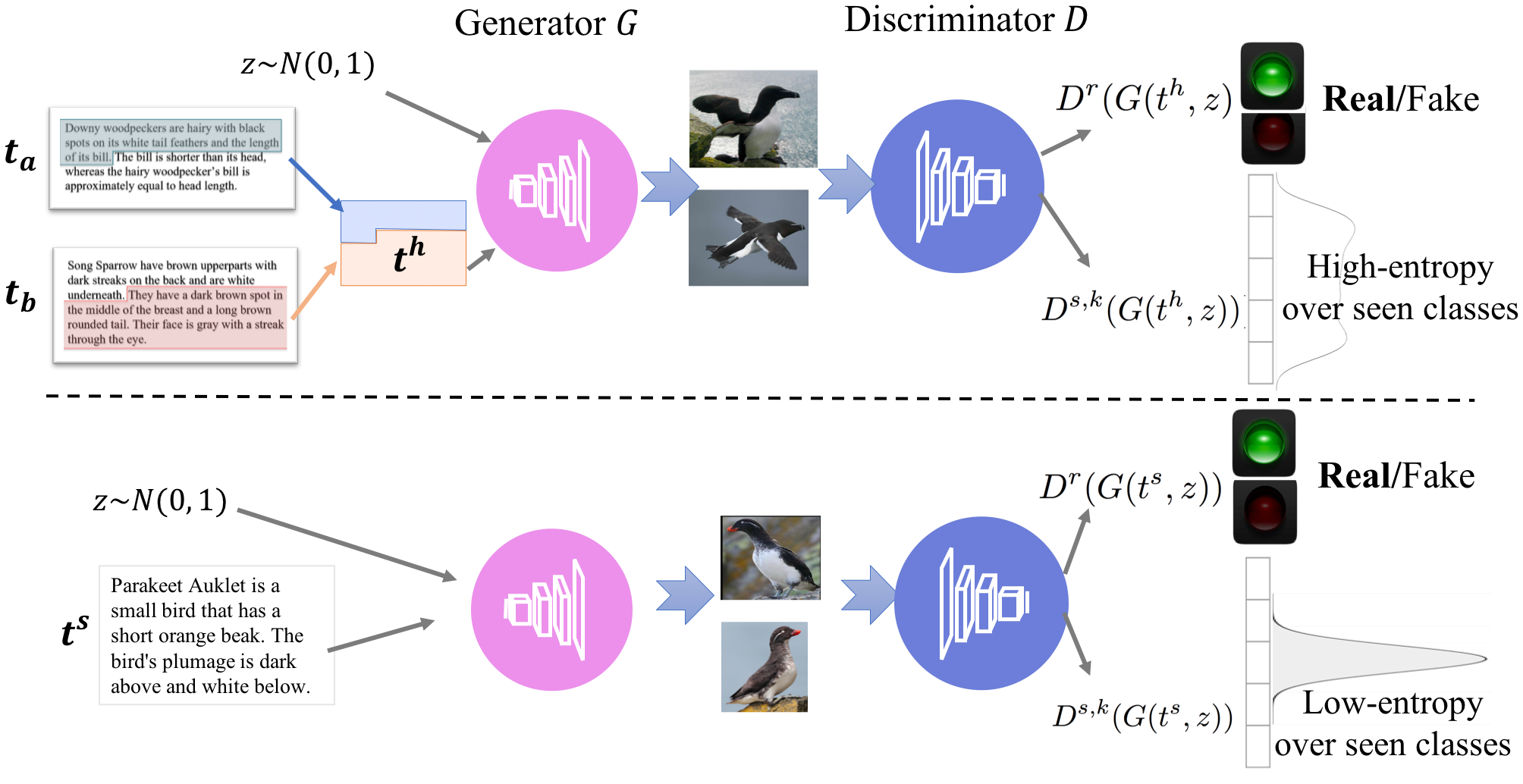}
	\vspace{-3mm}
	\caption{Generator $G$ is trained to carefully deviate from seen to unseen classes without synthesizing unrealstic images.  Top part: $G$ is provided with a hallucinated text $t^h$ and trained to trick discriminator to believe it is real, yet it encourages to deviate learning from seen classes by maximizing entropy over seen classes given $t^h$. Bottom part: $G$ is provided with text of a seen class $t^s$ and is trained to trick discriminator to believe it is real with a corresponding class label(low-entropy).}
	\label{fig:framework}
    \vspace{-5mm}
\end{figure*}
\noindent \textbf{Problem Definition.} We start by defining the zero-shot learning setting. 
We denote the semantic representations of unseen classes and seen classes as $t_i^u=\phi(T_k^u) \in \mathcal{T}$ and $t_i^s \in \mathcal{T}$ respectively, where $\mathcal{T}$ is the semantic space (e.g., features $\phi(\cdot)$ of a Wikipedia article $T_k^u$).  Let's denote the seen data as $D^s = \{(x_i^s, y_i^s, t_i^s) \}^{N^s}_{i=1}$, where $N^s$ is the number of training(seen) image examples,  where $x_i^s \in \mathcal{X}$ denotes the visual features of the $i^{th}$ image in the visual space $\mathcal{X}$, $y_i^s$ is the corresponding category label.  We denote the number of unique seen class labels as $K^s$.  We denote the set of seen and unseen class labels as $\mathcal{S}$ and $\mathcal{U}$, where  the aforementioned $y_i^s \in \mathcal{S}$. Note that the seen and the unseen classes  are disjointed, i.e.,  $\mathcal{S} \cap \mathcal{U} = \emptyset$. 
For unseen classes, we are given their semantic representations, one per class, $\{t_i^u\}_{i=1}^{K^u}$, where $K^u$ is the number of unseen classes. The zero-shot learning (ZSL) task is to predict the label $y_u \in \mathcal{U}$ of an  unseen class visual example  $x^u \in \mathcal{X} $. In the  more challenging Generalized ZSL (GZSL),  the aim is to predict $y \in \mathcal{U} \cup \mathcal{S}$ given  $x$ that may belong to seen or unseen classes. 

\noindent  \textbf{Approach Overview.} Fig.~\ref{fig:framework} shows an overview of our Creativity Inspired Zero-Shot Learning model(CIZSL).  Our method builds on top of GANs~\cite{goodfellow2014generative} while conditioning on semantic representation from raw Wikipedia text describing unseen classes. We denote the generator as $G$:  $\mathbb{R}^Z \times \mathbb{R}^T \xrightarrow{\theta_G} \mathbb{R}^X$ and  the discriminator as $D$ : $\mathbb{R}^X \xrightarrow{\theta_D} \{0,1\} \times \mathbb{L}_{cls}$, where  $\theta_G$ and $\theta_D$ are parameters of the generator and the discriminator as  respectively, $\mathbb{L}_{cls}$ is the set of seen class labels (i.e., $\mathcal{S} = \{ 1 \cdots K^s\}$). For the Generator $G$ and as in ~\cite{xian2018feature}, the text representation is then concatenated with a random vector $z \in \mathbb{R}^Z$ sampled from Gaussian distribution $\mathcal{N} (0,1)$; see Fig.~\ref{fig:framework}.  In the architecture of ~\cite{Elhoseiny_2018_CVPR}, the encoded text $t_k$ is first fed to a fully connected layer to reduce the dimensionality and to  suppress the noise  before concatenation with $z$. 
In our work, the discriminator $D$ is trained not only to predict real for images from the training images and fake for generated ones, but also to identify the category of the input image.  
We denote the  real/fake probability produced by $D$ for an input image as $D^r(\cdot)$, and the classification score of a seen class $k \in \mathcal{S}$  given the image as $D^{s,k}(\cdot)$. Hence, the features are generated from the encoded text description $t_k$, as follows $ \tilde{x}_k \leftarrow G(t_k, z)$. The discriminator then has two heads. The first head is an FC layer that for binary real/fake classification. The second head is a $K^s$-way classifier over the seen classes. 
Once our generator is trained, it is then used to hallucinate fake generations for unseen classes, where conventional classifier could be trained as we detail later in Sec~\ref{prediction}.

The generator $G$ is the key imagination component that we aim to train to generalize to unseen classes guided by signals from the discriminator $D$. In Sec~\ref{sec_cizsl}, we detail the definition of our Creativity Inspired Zero-shot Signal to augment and improve the learning capability of the Generator $G$. In Sec~\ref{creativegeneration}, we show how our proposed loss can be easily integrated into adversarial generative training.

\vspace{-1mm}
\subsection{Creativity  Inspired  Zero-Shot  Loss (CIZSL) }
\label{sec_cizsl}
\vspace{-1mm}
We explicitly explore the unseen/creative space of the generator $G$ with a  hallucinated text ($t^h\sim p^h_{text}$). We define $p^h_{text}$, as a probability distribution over hallucinated text description that is likely to be unseen and hard negatives to seen classes. To sample $t^h \sim p^h_{text}$, we first pick two seen text features at random $t^s_a, t^s_b \in \mathcal{S}$. Then we sample $t^h$ by interpolating between them as
\begin{equation}
\small 
  t^h= \alpha t^s_a + (1-\alpha) t^s_b 
  \label{eq_th}
\end{equation}
where $\alpha$ is uniformally sampled between $0.2$ and $0.8$. We discard $\alpha$ values close to $0$ or $1$ to avoid sampling a text feature very close to a seen one.  We also tried different ways to sample $\alpha$ which modifies $p^h_{text}$ like fixed $\alpha=0.5$ or $\alpha\sim \mathcal{N}(\mu=0.5, \sigma=0.5/3)$ but we found uniformally sampling from 0.2 to 0.8 is simple yet effective; see ablations at Appendix E.
We define our \emph{creativity inspired zero-shot loss }$L_G^C$ based on $G(t^h,z)$ as follows
\begin{equation}
\begin{split}
    L_G^C = - \mathbb{E}_{z \sim {p_{z}, t^h \sim {p^s_{text}}}}[D^r(G(t^h, z))] \,\,\,\,\,\,\,\,\,\,\,\,\,\,\,\, \\ + \mathcolor{black}{\lambda \mathbb{E}_{z \sim {p_{z}, t^h \sim {p^h_{text}}}} [ L_{e}( \{ D^{s,k}(G({t^h}, z))\}_{ k=1 \to K^s })]}  
\end{split}
\label{eq_lgc}
\end{equation}
We encourage $G(t^h,z)$ to be real (first term) yet hard to classify to any of the seen classes (second term) and hence achieve more discrimination against seen classes; see Fig.~\ref{fig:framework} (top). More concretely, the first term encourage the generations given $t^h \sim p^h_{text}$ to trick the discriminator to believe it is real (i.e., maximize $D^r(G(t^h, z)$). This loss encourages the generated examples to stay realistic while deviating from seen classes. In the second term, we quantify the difficulty of classification by maximizing an entropy function $L_{e}$ that we define later in this section. Minimizing $L_G^C$ connects to the principal of least effort by  Martindale et.al. 1990, where exaggerated novelty would decrease the transferability from seen classes (see visualized in Fig.~\ref{fig:intro}). Promoting the aforementioned high entropy distribution incents discriminative generation. However, it does not disable knowledge transfer from seen classes since the unseen generations are encouraged to be an entropic combination of seen  classes. 
We did not model deviation from seen classes as an additional class with label $K^s+1$ that we always classify $G(t^h,z)$ to, since this reduces the knowledge transfer from seen classes as we demonstrate in our results. 

\noindent \textbf{Definition of $L_e$ :} 
$L_e$ is defined over the seen classes' probabilities, produced by the second discriminator head $\{D^{s,k}(\cdot)\}_{k=1 \to K^s}$ (i.e.,  the softmax output over the seen classes). We tried different entropy maximization losses. They are based on minimizing the divergence between the softmax distribution produced by the discriminator given the hallucinated text features and the uniform distribution. Concretely, the divergence, also known as relative entropy, is minimized between $\{ D^{s,k}(G({t^h}, z))\}_{ k=1 \to K^s }))$ and $\{\frac{1}{K^s}\}_{ k=1 \to K^s }$; see Eq~\ref{le_eq}. 
Note that similar losses has been studied in the context of creative visual generation of art and fashion(e.g., ~\cite{elgammal2017can,sbai2018design}). However, the focus there was mainly unconditional generation and there was no need to hallucinate the input text  $t^h$ to the generator, which is necessary in our case; see Sec~\ref{sec_bg}. In contrast, our work also relates  two different modalities (i.e., Wikipedia text and  images).

\begin{equation}
\small 
\begin{aligned}
L^{KL}_{e} = \sum_{k=1}^{K^s} \frac{1}{K^s} {{D^{s,k}(G(t^h,z))}} \,\,\,\,\,\,\,\,\,\,\,\,\,\,\,\,\,\,\,\,\,\,\,\,\,\,\,\,\,\,\,\,\,\,\,\,\,\,\,\,\,\,\,\,\,\,\,\,\,\,\,\,\,\,\,\,\,\,\,\,\,\,\,\,\,\,\,\,\,\,\,\,\,\,\,\,\,\,\,\\
L^{SM}_{e}(\gamma,\beta) =\frac{1}{\beta-1} \left[ \sum_{k=1}^{K^s} (D^{s,k}(G(t^h,z)^{1-\gamma} {(\frac{1}{K^s})}^{\gamma})^\frac{1-\beta}{1-\gamma} -1\right]
\end{aligned}
\label{le_eq}
\end{equation}

Several divergence/entropy measures has been proposed in the information theory literature~\cite{Renyi60,Tsallis88,Bhattacharyya,IsSM07,sm_1976}. We adopted two divergence losses, the well-known Kullback-Leibler(KL) divergence in $L^{KL}_{e}$ and the two-parameter Sharma-Mittal(SM)~\cite{sm_1976} divergence in $L^{SM}_{e}$ which is relatively less known; see Eq~\ref{le_eq}. It was shown in~\cite{IsSM07},  that other divergence measures are special case of Sharma-Mittal(SM) divergence by setting its two parameters $\gamma$ and $\beta$. It is equivalent to  \emph{R\'enyi}~\cite{Renyi60} when $\beta \to 1$ (single-parameter), \emph{Tsallis} divergence~\cite{Tsallis88} when $\gamma=\beta$ (single-parameter), Bhattacharyya divergence when$\beta \to 0.5$ and $\gamma \to 0.5$,  and \emph{KL} divergence when $\beta \to 1$ and $\gamma \to 1$ (no-parameter). So, when we implement SM  loss, we can also  minimize any of the aforementioned special-case  measures; see details in Appendix B. Note that we also learn $\gamma$ and $\beta$ when we train our model with SM loss.

\subsection{Integrating CIZSL in Adversarial Training} 
\label{creativegeneration}
The integration of our approach is simple that $L_G^C$ defined in Eq~\ref{eq_lgc} is just added to the generator loss; see Eq~\ref{eq:gen}. Similar to existing methods, when the generator $G$ is provided with  text describing a seen class $t^s$, its is trained to trick the discriminator to believe it is real and to predict the corresponding class label (low-entropy for  $t^s$ versus hig-entropy for $t^h$); see Fig~\ref{fig:framework}(bottom). Note that the remaining terms, that we detail here for concreteness of our method, are simlar to existing generative ZSL approaches~\cite{xian2018feature,Elhoseiny_2018_CVPR} 

\noindent\textbf{Generator Loss}
\label{sec_gen_loss}
The generator loss is an addition of four terms, defined as follows
\vspace{-0.5em}
\begin{equation}
\small 
\begin{aligned}
L_G = { L_G^C} - \mathbb{E}_{z \sim {p_{z}, (t^s,y^s) \sim {p^s_{text}}}}[D^r(G(t^s, z))+ \,\,\,\,\,\,\,\,\,\,\,\,\,\,\,\,\,\,\,\,\,\,\,\,\,\,\,\, \,\,\,\,\,\,\,\,\\ \sum_{k=1}^{K^s} y^s_k log(D^{s,k}(G(t^s, z)))]  \\
+ \frac{1}{K^s}\sum_{k=1}^{K^s}|| \mathbb{E}_{z \sim {p_{z}}}[G(t_k, z)] - \mathbb{E}_{x \sim {p^k_{data}}}[x]||^2 \,\,\,\,\,\,\,\,\,\,\,\,
\label{eq:gen}
\end{aligned}
\end{equation}
The first term is our creativity inspired zero-shot loss $L_G^C$, described in Sec~\ref{sec_cizsl}. 
Note that seen class text descriptions $\{t_k\}_{k=1 \to K^s}$ are encouraged to predict a low entropy distribution since loss is minimized when the corresponding class is predicted with a high probability. Hence, the second term tricks the generator to classify visual generations from seen text $t^s$ as real. The third term encourages the generator to be capable of generating visual features conditioned on a given seen text.  The fourth term is an additional visual pivot regularizer that we adopted from~\cite{Elhoseiny_2018_CVPR}, which encourages the centers of the generated (fake) examples for each class $k$ (i.e., with $G(t_k, z)$) to be close to the centers of real ones from sampled from $p^k_{data}$ for the same class $k$. 
Similar to existing methods, the loss for the discriminator is defined as:
\begin{equation}
\small 
\begin{aligned}
L_D = & \mathbb{E}_{z \sim {p_{z}}, (t^s,y^s) \sim {p^s_{text}}}[D^r(G(t^s, z))] - \mathbb{E}_{x \sim {p_{data}}}[D^r(x)]  \\
&+  L_{Lip} -  \frac{1}{2}  \mathbb{E}_{x,y \sim {p_{data}}}[\sum_{k=1}^{K^s} y_k log(D^{s,k}(x)) ] \\
&-  \frac{1}{2} \mathbb{E}_{z\sim p_z, (t^s,y^s) \sim {p^s_{text}}}[\sum_{k=1}^{K^s} y^s_k log(D^{s,k}(G(t^s, z))) ] \\ 
\end{aligned}
\label{eq:disc}
\end{equation}
where $y$ is a one-hot vector encoding of the seen class label for the sampled image $x$, $t^s$ and $y^s$ are
features of a text description and the corresponding on-hot label sampled from seen classes $p^s_{text}$. The first two terms approximate Wasserstein distance of the distribution of real features and fake features. The third term is the gradient penalty to enforce the Lipschitz constraint: $L_{Lip} =  (||\bigtriangledown_{\tilde{x}} D^r(\tilde{x})||_2 - 1)^2$, where  $\tilde{x}$ is the linear interpolation of the real feature $x$ and the fake feature $\hat{x}$; see~\cite{gulrajani2017improved}.  The last two terms are classification losses of the seen real features and fake features from  text descriptions of seen category labels.
\noindent \textbf{Training.}  We construct two minibatches for training the generator $G$, one from seen class $t^s$ and from the halllucinated text $t^h$ to minimize $L_G$ (Eq.~\ref{eq:gen}) and in particular $L_G^C$ (Eq.~\ref{eq_lgc}).  The generator is optimized to fool the discriminator into believing the generated features as real  either from hallucinated text $t^h$ or the seen text  $t^s$. In the mean time,  we maximize their entropy over the seen classes if the generated features comes from hallucinated text $t^h \sim p^h_{text}$ or to the corresponding class if from a real text $t^s$. Training the discriminator is similar to existing works;  see in Appendix C  a detailed algorithm and code to show how $G$ and $D$ are alternatively trained with an Adam optimizer. Note that when $L_e$ has parameters like $\gamma$ and $\beta$ for Sharma-Mittal(SM) divergence ( Eq~\ref{le_eq}), that we also learn. 

\subsection{Zero-Shot Recognition Test}
\label{prediction}
After training, the visual features of unseen classes can be synthesized by the generator conditioned on a given unseen text  description $t_u$, as $x_u = G(t_u, z)$. 
We can generate an arbitrary number of generated visual features by sampling different $z$ for the same text $t_u$.
With this synthesized data of unseen classes, the zero-shot recognition becomes a conventional classification problem.  We  used nearest neighbor prediction, which we found simple and  effective.

\vspace{-2mm}
\section{Experiments}


We investigate the performance of our approach on two class-level semantic settings: textual  and attribute descriptions. Since the textual based ZSL is a harder problem, we used it to run an ablation study for  zero-shot retrieval and generalized ZSL. Then, we conducted  experiments for both settings to validate the generality of our work.

\noindent \textbf{Cross-Validation} 
The weight $\lambda$ of our loss in Eq~\ref{eq_lgc} is a hyperparameter  that we found easy to tune on all of our experiments. 
We start by splitting the data into training and validation split with nearly 80-20\% ratio for all settings. Training and validation classes are selected randomly prior to the training. Then, we compute validation performance when training the model on the 80\% split every 100 iterations out of 3000 iterations. We investigate a wide range of values for $\lambda$, and the value that scores highest validation performance is selected to be used at the inference time. Finally, we combine training and validation data and evaluate the performance on testing data.

\noindent \textbf{Zer-Shot Performance Metrics.} We use two metrics widely used in in evaluating ZSL recognition performance: Standard Zero-shot recognition with the Top-1 unseen class accuracy and Seen-Unseen Generalized Zero-shot performance with Area under Seen-Unseen curve~\cite{chao2016empirical}. The Top-1 accuracy is the average percentage   of  images from unseen classes classifying correctly to one of unseen class labels. However, this might be  incomplete measure since it is more realistic at inference time to encounter also seen classes. Therefore, We also report a generalized zero-shot recognition metric with respect to the seen-unseen  curve, proposed by Chao~\emph{et al.}~\cite{chao2016empirical}. This metric classifies images of both seen $\mathcal{S}$ and unseen classes $\mathcal{U}$ at test time. Then, the performance of a ZSL model is assessed by classifying these images to the label space that covers both seen classes and unseen labels $\mathcal{T} = \mathcal{S}\cup\mathcal{U}$. A balancing parameter is used sample seen and unseen class test accuracy-pair. This pair is plotted as the $(x,y)$ co-ordinate to form the Seen-Unseen Curve(SUC). We follow ~\cite{Elhoseiny_2018_CVPR} in using the Area Under SUC to evaluate the generalization capability of class-level text zero-shot recognition, and the haromnic mean of SUC for attribute-based zero-shot recognition. In our model, we use the trained GAN to synthesize the visual features for both training and testing classes.

\begin{table*}
\vspace{-4mm}
\parbox{.60\linewidth}{
\centering
\resizebox{.60\textwidth}{!}{%
\begin{tabular}{@{}cccccccccc@{}}
\toprule
Metric & \multicolumn{4}{c}{Top-1 Accuracy (\%)} &  & \multicolumn{4}{c}{Seen-Unseen AUC (\%)} \\ \cmidrule(lr){2-5} \cmidrule(l){7-10} 
Dataset & \multicolumn{2}{c}{CUB} & \multicolumn{2}{c}{NAB} &  & \multicolumn{2}{c}{CUB} & \multicolumn{2}{c}{NAB} \\
Split-Mode & Easy & Hard & Easy & Hard &  & Easy & Hard & Easy & Hard \\ \midrule
\textbf{CIZSL SM-Entropy (ours final)} & \textbf{44.6} & \textbf{14.4} & 36.5 & \textbf{9.3} &  & \textbf{39.2} & \textbf{11.9} & \textbf{24.5} & \textbf{6.4} \\ \hline 
CIZSL SM-Entropy (replace $2^{nd}$ term in Eq~\ref{eq_lgc} by Classifying $t^h$ as new class) & 43.2 & 11.31 & 35.6 & 8.5 &  &38.3 & 9.5 & 21.6 & 5.6\\ 
CIZSL SM-Entropy (minus  $1^{st}$ term in Eq~\ref{eq_lgc}) & 43.4 & 10.1 & 35.2 & 8.3 &  & 35.0 & 8.2 & 20.1  & 5.4\\  
CIZSL SM-Entropy: (minus  $2^{nd}$ term in Eq~\ref{eq_lgc}) & 41.7 & 11.2 & 33.4 & 8.1 &  & 33.3 & 10.1 & 21.3 & 5.1\\
\hline 
CIZSL Bachatera-Entropy ($\gamma=0.5,\beta =0.5$) & 44.1 & 13.7 & 35.9 & 8.9 &  & 38.9 & 10.3 & 24.3 & 6.2\\
CIZSL Renyi-Entropy ($\beta \to 1$) &  44.1 & 13.3 & 35.8 & 8.8 &  & 38.6 & 10.3 & 23.7 & 6.3\\	
CIZSL KL-Entropy ($\gamma \to 1,\beta \to 1$)	&  44.5 & 14.2 & 36.3 & 8.9 &  & 38.9 & 11.6 & 24.3 & 6.2\\
CIZSL Tsallis-Entropy ($\beta = \gamma$) & 44.1 & 13.8 & \textbf{36.7} & 8.9 &  & 38.9 & 11.3 & 24.5 & 6.3  \\	  \hline
CIZSL SM-Entropy: (minus  $1^{st}$ and $2^{nd}$ terms in Eq~\ref{eq_lgc})= GAZSL~\cite{Elhoseiny_2018_CVPR}& 43.7 & 10.3 & 35.6 & 8.6 &  & 35.4 & 8.7 & 20.4 & 5.8 \\ 
\bottomrule
\end{tabular}%
}
\caption{Ablation Study using Zero-Shot recognition on \textbf{CUB} \& \textbf{NAB} datasets with two split settings each. CIZSL is GAZSL~\cite{Elhoseiny_2018_CVPR}+ our loss}
\label{tb:nab_cub_ablation}
}
\hfill
\parbox{.38\linewidth}{
\centering
\includegraphics[width=0.30\textwidth, height=3.62cm]{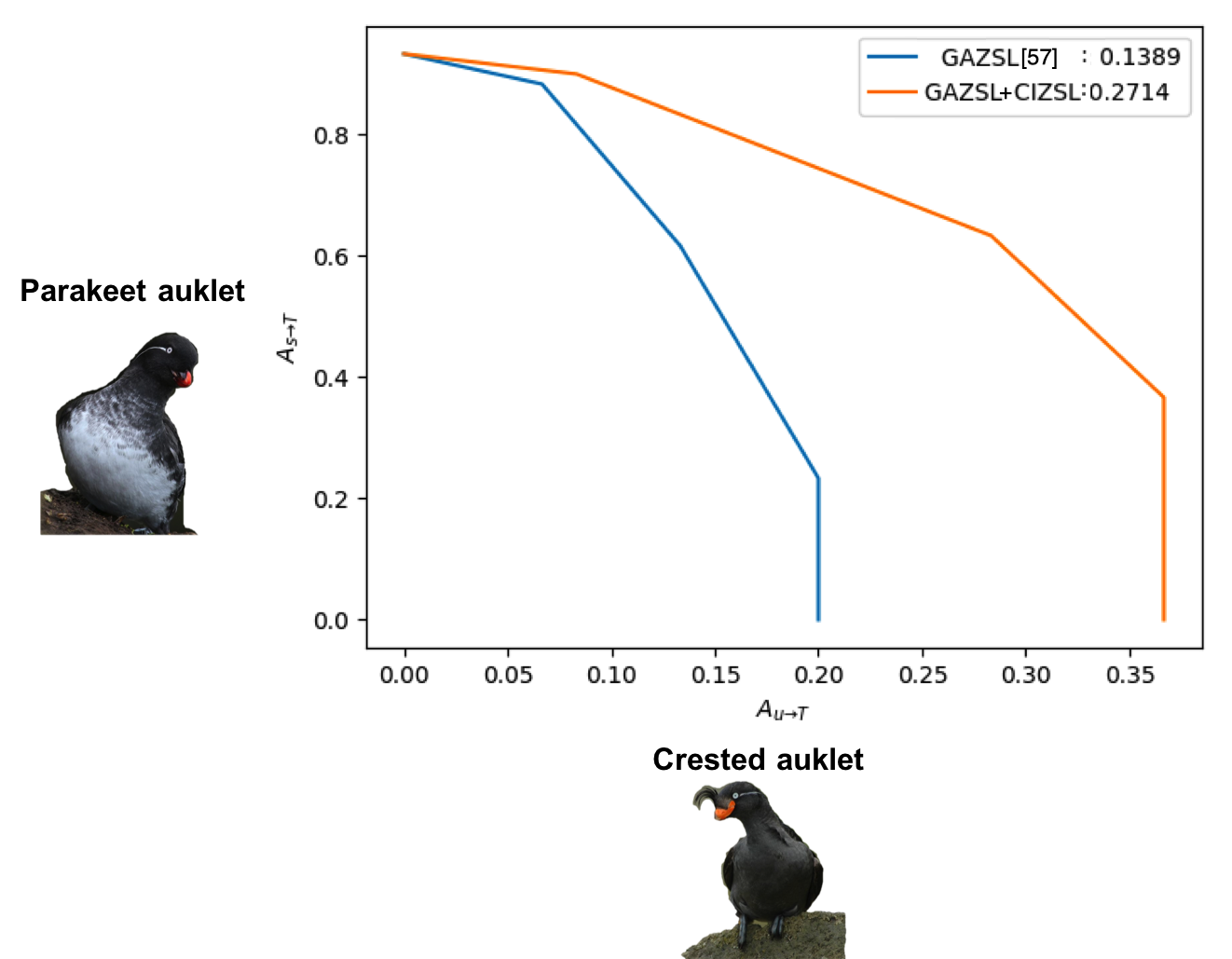}
		\vspace{-3mm}
\captionof{figure}{Seen Unseen Curve for Parakeet Auklet (Seen,  y-axis) vs Crested Auklet (Unseen, x-axis) for GAZSL[57] and     GAZSL[57]+ CIZSL.}
\label{fig_auklet}
}
\vspace{-3.67ex}
\end{table*}

\subsection{Wikipedia based ZSL  Results (4 benchmarks)}
\noindent \textbf{Text Representation.} Textual features for each class are extracted from corresponding raw Wikipedia articles collected by~\cite{elhoseiny2013write,Elhoseiny_2017_CVPR}. 
We used Term Frequency-Inverse Document Frequency (TF-IDF)~\cite{salton1988term} feature vector of dimensionality 7551 for CUB and 13217 for NAB.  

\noindent \textbf{Visual Representation.} We use features of the part-based FC layer in VPDE-net~\cite{zhang2016learning}. The image are fed forward to the VPDE-net after resizing to $224\times 224$, and the feature activation for each detected part is extracted which is of 512 dimensionality. The dimensionalities of visual features for CUB and NAB are 3583 and 3072 respectively. There are six semantic parts shared in CUB and NAB:``head", ``back", ``belly", ``breast", ``leg", ``wing", ``tail". Additionally, CUB has an extra part which is ``leg" which makes its feature representation 512D longer compared to NAB (3583 vs 3072). More details in the Appendix F.

\noindent \textbf{Datasets.} We use two common fine-grained recognition datasets for textual descriptions: \emph{Caltech UCSD Birds-2011} (CUB)~\cite{wah2011caltech} and \emph{North America Birds} (NAB)~\cite{Horn2015}. CUB dataset contains 200 classes of bird species and their Wikipedia textual description constituting a total of 11,788 images. Compared to CUB, NAB is a larger dataset of birds, containing a 1011 classes and 48,562 images.

\noindent \textbf{Splits.} For both datasets, there are two schemes to split the classes into training/testing (in total four benchmarks): Super-Category-Shared (SCS) or \textit{easy} split and Super-Category-Exclusive Splitting (SCE) or \textit{hard} split, proposed in~\cite{Elhoseiny_2017_CVPR}. Those splits represents the similarity of the seen to unseen classes, such that the former represents a higher similarity than the latter. For SCS (easy), unseen classes are deliberately picked such that for every unseen class, there is at least one seen class with the same super-category. Hence, the relevance between seen and unseen classes is very high, deeming the zero-shot recognition and retrieval problems relatively easier. On the other end of the spectrum,  SCE (hard) scheme, the unseen classes do not share the super-categories with the seen classes. Hence, there is lower similarity between the seen and unseen classes making the problem harder to solve. Note that the easy split is more common in literature since it is more Natural yet the deliberately designed hard-split shows the progress when the super category is not seen that we also may expect. 

\begin{table*}
\parbox{.57\linewidth}{
\centering
\resizebox{.56\textwidth}{!}{%
\begin{tabular}{@{}cccccccccc@{}}
\toprule
Metric & \multicolumn{4}{c}{Top-1 Accuracy (\%)} &  & \multicolumn{4}{c}{Seen-Unseen AUC (\%)} \\ \cmidrule(lr){2-5} \cmidrule(l){7-10} 
Dataset & \multicolumn{2}{c}{CUB} & \multicolumn{2}{c}{NAB} &  & \multicolumn{2}{c}{CUB} & \multicolumn{2}{c}{NAB} \\
Split-Mode & Easy & Hard & Easy & Hard &  & Easy & Hard & Easy & Hard \\ \midrule
WAC-Linear \cite{elhoseiny2013write} & 27.0 & 5.0 & -- & -- &  & 23.9 & 4.9 & 23.5 & -- \\
WAC-Kernel \cite{elhoseiny2016write} & 33.5 & 7.7 & 11.4 & 6.0 &  & 14.7 & 4.4 & 9.3 & 2.3 \\
ESZSL \cite{romera2015embarrassingly} & 28.5 & 7.4 & 24.3 & 6.3 &  & 18.5 & 4.5 & 9.2 & 2.9 \\
ZSLNS \cite{Qiao2016} & 29.1 & 7.3 & 24.5 & 6.8 &  & 14.7 & 4.4 & 9.3 & 2.3 \\
SynC$_{fast}$ \cite{changpinyo2016synthesized} & 28.0 & 8.6 & 18.4 & 3.8 &  & 13.1 & 4.0 & 2.7 & 3.5 \\
ZSLPP \cite{Elhoseiny_2017_CVPR} &  37.2 & 9.7 &30.3 & 8.1 &  & 30.4 & 6.1 & 12.6 & 3.5 \\ \hline
FeatGen~\cite{xian2018feature} &  43.9 & 9.8 &36.2 & 8.7 &  & 34.1 & 7.4 & 21.3 & 5.6 \\
FeatGen\cite{xian2018feature}+ \textbf{CIZSL} &  44.2$\mathcolor{blue}{\mathbf{ ^{+0.3}}}$  & 12.1 $\mathcolor{blue}{\mathbf{ ^{+2.3}}}$  & 36.3 $ \mathcolor{blue}{\mathbf{ ^{+0.1}}}$ & \textbf{9.8} $\mathcolor{blue}{\mathbf{ ^{+1.1}}}$ &  & 37.4 $\mathcolor{blue}{\mathbf{ ^{+2.7}}}$ & 9.8$\mathcolor{blue}{\mathbf{ ^{+2.4}}}$ & 24.7$\mathcolor{blue}{\mathbf{ ^{+3.4}}}$ & 6.2 $\mathcolor{blue}{\mathbf{ ^{+0.6}}}$ \\  \hline
GAZSL~\cite{Elhoseiny_2018_CVPR}& 43.7 & 10.3 & 35.6 & 8.6 &  & 35.4 & 8.7 & 20.4 & 5.8 \\    \midrule
GAZSL~\cite{Elhoseiny_2018_CVPR} + \textbf{CIZSL} &  \textbf{44.6} $\mathcolor{blue}{\mathbf{ ^{+0.9}}}$ & \textbf{14.4} $\mathcolor{blue}{\mathbf{ ^{+4.1}}}$ & \textbf{36.6} $\mathcolor{blue}{\mathbf{ ^{+1.0}}}$ & 9.3 $\mathcolor{blue}{\mathbf{ ^{+0.7}}}$ &  & \textbf{39.2}$\mathcolor{blue}{\mathbf{ ^{+3.8}}}$ & \textbf{11.9}$\mathcolor{blue}{\mathbf{ ^{+3.2}}}$ & \textbf{24.5}$\mathcolor{blue}{\mathbf{ ^{+4.1}}}$ & \textbf{6.4}$\mathcolor{blue}{\mathbf{ ^{+0.6}}}$\\ \bottomrule
\end{tabular}%
}
\caption{Zero-Shot Recognition on class-level textual description from \textbf{CUB} and \textbf{NAB} datasets with two-split setting.}
\label{tb:nab_cub}
\vspace{-1.67ex}
}
\hfill
\parbox{.4\linewidth}{
\centering
\resizebox{.4\textwidth}{!}{%
\begin{tabular}{@{}cccccccc@{}}
\toprule
 & \multicolumn{3}{c}{Top-1 Accuracy(\%)} &  & \multicolumn{3}{c}{Seen-Unseen H} \\ \cmidrule(lr){2-4} \cmidrule(l){6-8} 
 & AwA2 & aPY & SUN &  & AwA2 & aPY & SUN \\ \midrule
DAP~\cite{Lampert2014} & 46.1 & 33.8 & 39.9 &  & -- & 9.0 & 7.2 \\
SSE~\cite{zhang2015zero} & 61.0 & 34.0 & 51.5 &  & 14.8 & 0.4 & 4.0 \\
SJE~\cite{akata2015evaluation} & 61.9 & 35.2 & 53.7 &  & 14.4 & 6.9 & 19.8 \\
LATEM~\cite{xian2016latent} & 55.8 & 35.2 & 55.3 &  & 20.0 & 0.2 & 19.5 \\
ESZSL~\cite{romera2015embarrassingly} & 58.6 & 38.3 & 54.5 &  & 11.0 & 4.6 & 15.8 \\
ALE~\cite{akata2016label} & 62.5 & 39.7 & 58.1 &  & 23.9 & 8.7 & 26.3 \\
CONSE~\cite{norouzi2013zero} & 44.5 & 26.9 & 38.8 &  & 1.0 & -- & 11.6 \\
SYNC~\cite{changpinyo2016synthesized} & 46.6 & 23.9 & 56.3 &  & 18.0 & 13.3 & 13.4 \\
SAE~\cite{kodirov2017semantic} & 54.1 & 8.3 & 40.3 &  & 2.2 & 0.9 & 11.8 \\
DEM~\cite{zhang2016learning} & 67.1 & 35.0 & 61.9 &  & 25.1 & 19.4 & 25.6 \\
DEVISE~\cite{frome2013devise} & 59.7 & 39.8 & 56.5 &  & \textbf{27.8} & 9.2 & 20.9 \\  \hline
GAZSL~\cite{Elhoseiny_2018_CVPR} & 58.9 & 41.1 & 61.3 &  & 15.4 & 24.0 & 26.7 \\ 
GAZSL~\cite{Elhoseiny_2018_CVPR} + \textbf{CIZSL} & \textbf{67.8} $\mathcolor{blue}{\mathbf{ ^{+8.9}}}$ & {42.1} $\mathcolor{blue}{\mathbf{ ^{+1.0}}}$ & {63.7} $\mathcolor{blue}{\mathbf{ ^{+2.4}}}$ &  & $24.6 \mathcolor{blue}{\mathbf{ ^{+9.2}}}$ & {25.7} $\mathcolor{blue}{\mathbf{ ^{+1.7}}}$ & \textbf{27.8} $\mathcolor{blue}{\mathbf{ ^{+1.1}}}$  \\
\hline \hline
FeatGen~\cite{xian2018feature} & 54.3 & 42.6 & 60.8 &  & 17.6 & 21.4 & 24.9 \\
FeatGen~\cite{xian2018feature} + \textbf{CIZSL} & $60.1\mathcolor{blue}{\mathbf{ ^{+5.8}}}$  & $43.8\mathcolor{blue}{\mathbf{ ^{+1.2}}}$ & $59.4\mathcolor{red}{\mathbf{ ^{-0.6}}}$ &  & $19.1\mathcolor{blue}{\mathbf{ ^{+1.5}}}$  & $24.0 \mathcolor{blue}{\mathbf{ ^{+2.6}}}$ & $26.5 \mathcolor{blue}{\mathbf{ ^{+1.6}}}$ \\
\midrule
cycle-(U)WGAN~\cite{felix2018multi} & 56.2 & 44.6 & 60.3 &  & 19.2 & 23.6 & 24.4 \\
cycle-(U)WGAN~\cite{felix2018multi} + \textbf{CIZSL} & $63.6 \mathcolor{blue}{\mathbf{ ^{+7.4}}}$  & \textbf{45.1} $\mathcolor{blue}{\mathbf{ ^{+0.5}}}$  & \textbf{64.2} $\mathcolor{blue}{\mathbf{ ^{+3.9}}}$   &  & $23.9 \mathcolor{blue}{\mathbf{ ^{+4.7}}}$   & \textbf{26.2} $\mathcolor{blue}{\mathbf{ ^{+2.6}}}$   & $27.6 \mathcolor{blue}{\mathbf{ ^{+3.2}}}$  \\
\bottomrule
\end{tabular}%
}
\caption{Zero-Shot Recognition on class-level attributes of \textbf{AwA2}, \textbf{aPY} and \textbf{SUN} datasets.}
\label{tb:awa2_apy_sun}
}
\vspace{-2.0ex}
\end{table*}
\noindent \textbf{Ablation Study (Table~\ref{tb:nab_cub_ablation}).}  Our loss is composed of two terms shown that encourage the careful  deviation in Eq~\ref{eq_lgc}.  The first term encourages that the generated visual features from the hallucinated text $t^h$ to deceive the discriminator believing it is real, which restricts synthesized visual features to be realistic. The second term maximizes the entropy using a deviation measure.  
In our work, Shama-Mittal(SM) entropy  parameters $\gamma$ and $\beta$ are learnt and hence adapt the corresponding data and split mode to a matching divergence function, leading to the best results especially in the generalized SUAUC metric; see first row in Table~\ref{tb:nab_cub_ablation}. We first investigate the effect of deviating the hallucinated text by classifying it t a new class $K^s+1$, where $K^s$ is the number of the seen classes. We found the performance is significantly worse since the loss would significantly increase indecencies against seen classes and hence reduces seen knowledge transfer to unseen classes; see row 2 in  Table~\ref{tb:nab_cub_ablation}. When we remove the first term (realistic constraints), the performance degrades  especially under the generalized Seen-Unseen AUC metric because generated visual features became unrealistic; see row 3 in  Table~\ref{tb:nab_cub_ablation} (e.g., 39.2\% to 35.0\% AUC drop for CUB Easy and 11.9\%-8.2\% drop for CUB Hard). 
Alternatively, when we remove the second term (entropy), we also observe a significant drop in performance showing that both losses are complementary to each other; see row 4 in  Table~\ref{tb:nab_cub_ablation} (e.g., 39.2\% to 33.5\% AUC drop for CUB Easy and 11.9\%-10.1\% drop for CUB Hard). 
In our ablation,  applying our approach without both terms (our loss) is equivalent to~\cite{Elhoseiny_2018_CVPR}, shown is the last row in Table~\ref{tb:nab_cub_ablation} as one of the least performing baselines. Note that our loss is applicable to other generative ZSL methods as we show in our state-of-the-art comparisons later in this section. 





We also compare different entropy measures to encourage the deviation from the seen classes: \emph{Kullback-Leibler (KL)}, \emph{R\'enyi}~\cite{Renyi60}, \emph{Tsallis}~\cite{Tsallis88}, \emph{Bhattacharyya}~\cite{Bhattacharyya}; see rows 5-8 in Table~\ref{tb:nab_cub_ablation}. All these divergences measure are special cases of  the two parameter ($\gamma$ , $\beta$) Sharma-Mittal(SM)~\cite{sm_1976} divergence that we implemented.
For instance, Renyi~\cite{Renyi60} and Tsallis~\cite{Tsallis88} on the other hand only learns one parameter and achieves comparable yet lower performance.  Bhattacharyya~\cite{Bhattacharyya} and KL have no learnable parameters an achieves lower performance compared to SM.


\begin{figure}
	\centering
	\subfloat[CUB with SCS (easy) split]{\includegraphics[width= 1.6in]{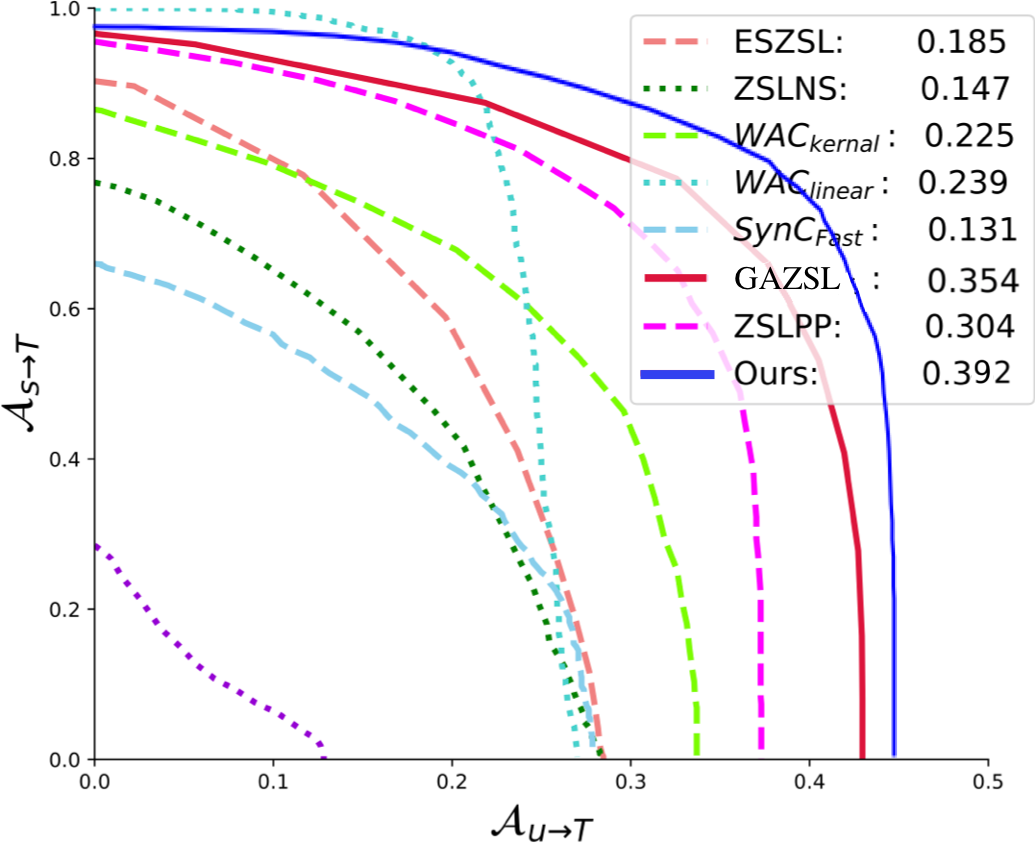}} 
	\subfloat[CUB with SCE (hard) split]{\includegraphics[width= 1.6in]{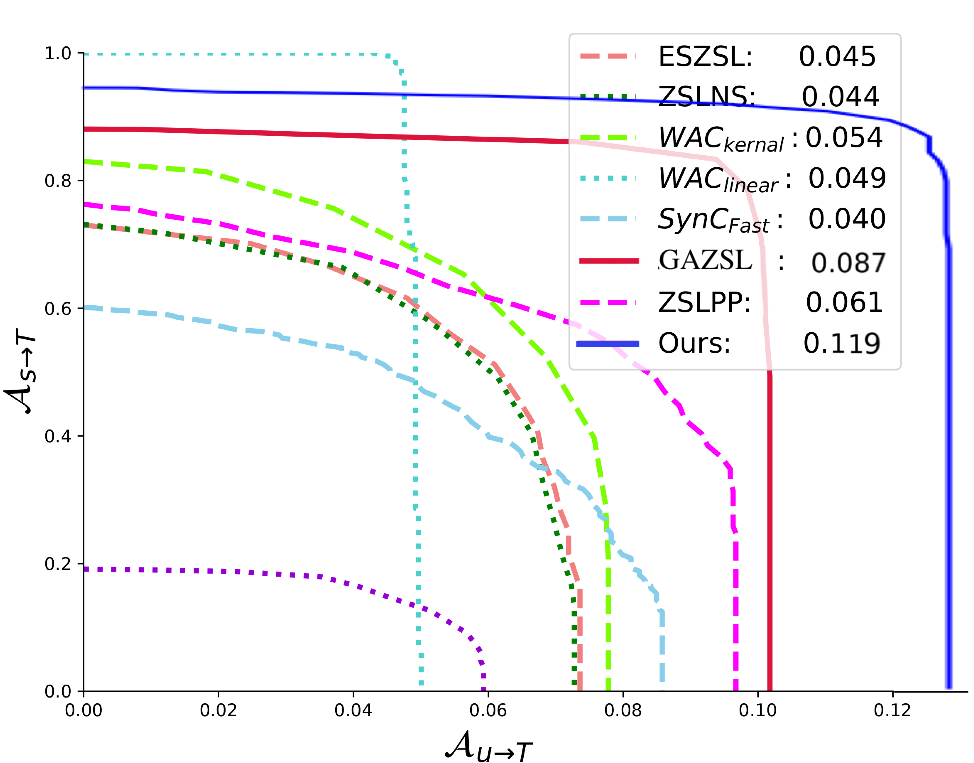}}\\
    \vspace{-2mm}
	\subfloat[NAB with SCS (easy) split]{\includegraphics[width= 1.6in]{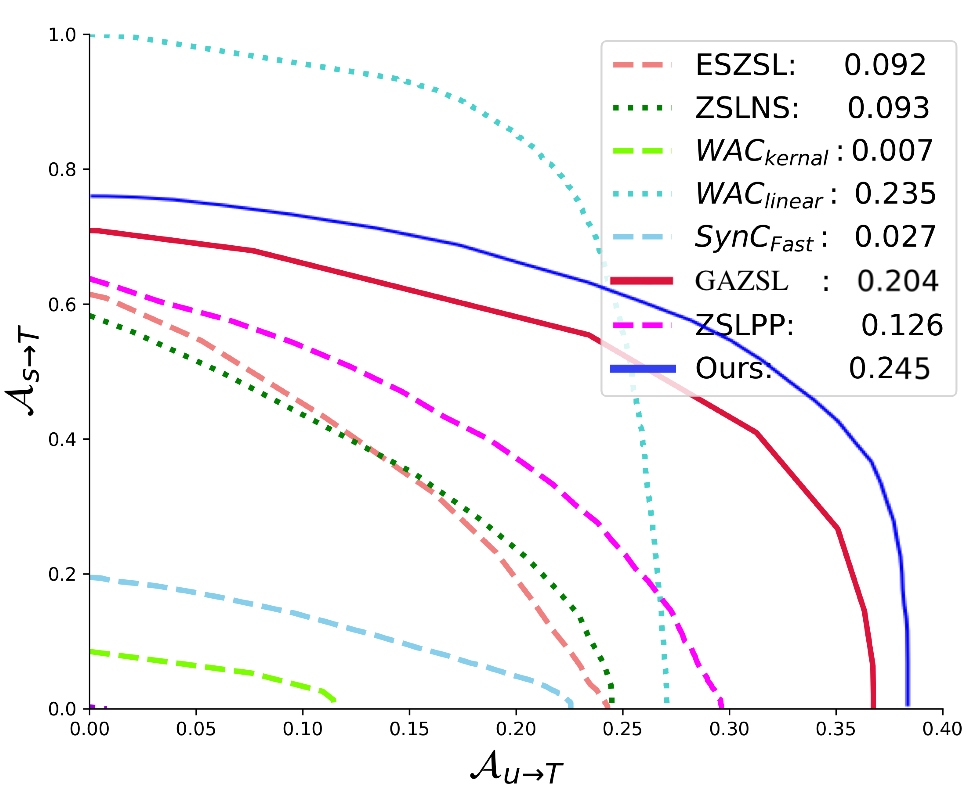}} 
	\subfloat[NAB with SCE (hard) split]{\includegraphics[width= 1.6in]{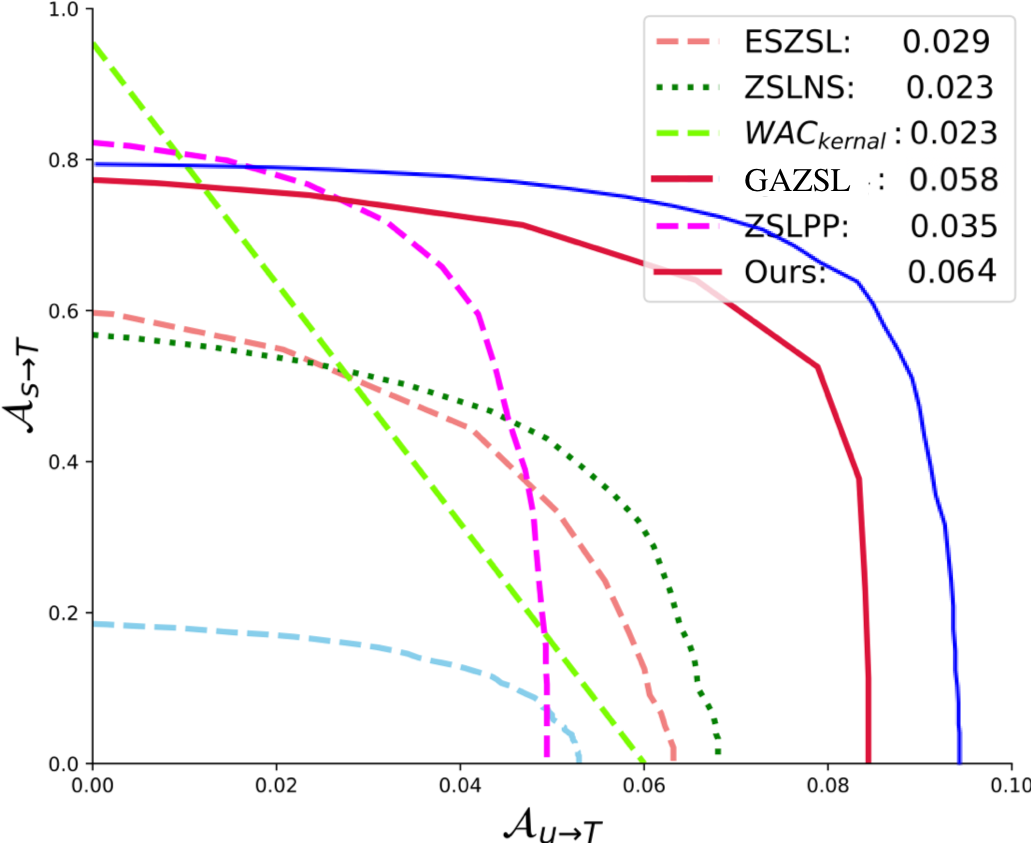}}\\
	\setlength\belowcaptionskip{-2.5ex}
	\caption{Seen-Unseen accuracy Curve with two splits: SCS(easy) and SCE(hard). Ours indicates GAZSL+ CIZSL }
	\label{fig:GZSL_Curve}
\end{figure}
%

\noindent \textbf{Zero-Shot Recognition and Generality on ~\cite{xian2018feature} and ~\cite{Elhoseiny_2018_CVPR}.}
Fig~\ref{fig_auklet} shows the key advantage of our CIZSL loss,  doubling the capability of [57] from 0.13 AUC to 0.27 AUC  to distinguish between two very similar birds: Parakeet Auklet (Seen class) and Crested Auklet (unseen class), in 200-way classification; see Appendix A for  details. Table~\ref{tb:nab_cub} shows state-of-the-art comparison on CUB and NAB datasets for both their SCS(easy) and SCE(hard) splits (total of four benchmarks). Our method shows a significant advantage compared to the state of the art especially in generalized Seen-Unseen AUC metric ranging from 1.0-4.5\% improvement. Fig~\ref{fig:GZSL_Curve} visualizes Seen-Unseen curves for our four benchmarks CUB (east and hard splits) and NABirds (easy and hard splits) where ours has a significant advantage compared to state-of-the-art on recognizing unseen classes; see our area under SU curve  gain in Fig~\ref{fig:GZSL_Curve}  against the runner-up GAZSL. The average relative SU-AUC improvement on the easy splits is 15.4\% and 23.56\% on the hard split. Meaning, the advantage of our loss becomes more clear as splits get harder, showing a better capability of discriminative knowledge transfer.  We show the generality of our method by embedding it with another feature generation method, FeatGen~\cite{xian2018feature}, causing a consistent improvement.  All the methods are using same text and visual representation.

\noindent \textbf{Zero-Shot Retrieval.} We investigate our model's performance for zero-shot retrieval task given the Wikipedia article of the class using mean Average Precision (mAP), the common retrieval metric. In table ~\ref{tb:Retrieval}, we report the performance of different settings: retrieving 25\%, 50\%, 100\% of the images at each class. 
We follow ~\cite{Elhoseiny_2018_CVPR} to obtain the visual center of unseen classes by generating 60 examples for the given text then computing the average. Thus, given the visual center, the aim is to retrieve images based on the nearest neighbor strategy in the visual features space. Our model is the best performing and improves the MAP (100\%) over the  runner-up (~\cite{Elhoseiny_2018_CVPR}) by 14.64\% and 9.61\% on CUB and NAB respectively. 
Even when the model fails to retrieve the exact unseen class, it tends to retrieve visually similar images; see qualitative examples in Appendix D.



 \begin{table}[]
\centering
\scalebox{0.6}
{
\begin{tabular}{cccccccc}
\hline
\multirow{2}{*}{} & \multicolumn{3}{c}{CUB} &  & \multicolumn{3}{c}{NAB} \\ \cline{2-4} \cline{6-8} 
 & 25\% & 50\% & {100\%} &  & 25\% & 50\% &{ {100\% }}\\ \hline
ESZSL \cite{romera2015embarrassingly} & 27.9 & 27.3 & 22.7 &  & 28.9 & 27.8 & 20.9 \\
ZSLNS \cite{Qiao2016} & 29.2 & 29.5 & 23.9 &  & 28.8 & 27.3 & 22.1 \\
ZSLPP \cite{Elhoseiny_2017_CVPR} & 42.3 & 42.0 & 36.3 &  & 36.9 & 35.7 & 31.3 \\
GAZSL \cite{Elhoseiny_2018_CVPR} & 49.7 & 48.3 & 40.3 &  & \textbf{41.6} & 37.8 & 31.0 \\ \hline
GAZSL \cite{Elhoseiny_2018_CVPR}+ CIZSL & \textbf{50.3}$\mathcolor{blue}{\mathbf{ ^{+0.6}}}$ & \textbf{48.9} $\mathcolor{blue}{\mathbf{ ^{+0.6}}}$ & \textbf{46.2}$\mathcolor{blue}{\mathbf{ ^{+5.9}}}$ &  & 41.0$\mathcolor{red}{\mathbf{ ^{-0.6}}}$& \textbf{40.2}$\mathcolor{blue}{\mathbf{ ^{+2.4}}}$ & \textbf{34.2}$\mathcolor{blue}{\mathbf{ ^{+3.2}}}$\\ \hline
\end{tabular}%
}
\caption{Zero-Shot Retrieval using mean Average Precision(mAP) (\%) on CUB and NAB with SCS(easy) splits.}
\vspace{-3.67ex}
\label{tb:Retrieval}
\end{table}

\subsection{Attribute-based Zero-Shot Learning}
\noindent \textbf{Datasets.}  Although it is not our focus, we also investigate the performance of our model's zero-shot recognition ability using different semantic representation. We follow the GBU setting~\cite{gbu}, where images are described by their attributes instead of textual describtion deeming the problem to be relatively easier than textual-description zero-shot learning. We evaluated our approach on the following datasets: Animals with Attributes (AwA2)~\cite{lampert2009}, aPascal/aYahoo objects(aPY)~\cite{farhadi2009describing} and the SUN scene attributes dataset~\cite{patterson2012sun}. They consist of images covering a variety of categories in different scopes: animals, objects and scenes respectively. AwA contains attribute-labelled classes but aPY and SUN datasets have their attribute signature calculated as the average of the instances belonging to each class. 

\noindent \textbf{Zero-Shot Recognition.} 
On AwA2, APY, and SUN datasets, we show in Table~\ref{tb:awa2_apy_sun} that our CIZSL loss improves three generative zero-shot learning models including GAZSL~\cite{Elhoseiny_2018_CVPR},  FeatGen~\cite{xian2018feature},  and  cycle-(U)WGAN~\cite{felix2018multi}.
The table also shows our comparison to the state-of-the-art where we mostly obtain a superior performance. Even when obtaining a slightly lower score than state-of-the-art on AWA2, our loss adds a 9.2\% Seen-Unseen H absolute improvement to the non-creative GAZSL~\cite{Elhoseiny_2018_CVPR}. We also evaluated our loss on \emph{CUB-T1(Attributes) benchmark~\cite{gbu}}, where the Seen-Unseen H for  GAZSL~\cite{Elhoseiny_2018_CVPR}] and GAZSL~\cite{Elhoseiny_2018_CVPR}+CIZSL are 55.8 and  57.4, respectively. 

\vspace{-2mm}
\section{Conclusion}

We draw an inspiration from the psychology of human creativity to improve the capability of unseen class imagination for zero-shot recognition. We adopted GANs to discrimnatively imagine visual features given a hallucinated text describing an unseen visual class. Thus, our generator learns to synthesize unseen classes from hallucinated texts. Our loss encourages deviating generations of unseen from seen classes by enforcing a high entropy on seen class classification while being realistic. Nonetheless, we ensure the realism of hallucinated text by synthesizing visual features similar to the seen classes to preserve knowledge transfer to unseen classes. Comprehensive evaluation on seven benchmarks shows a consistent improvement over the state-of-the-art for both zero-shot learning and retrieval with class description defined by  Wikipedia articles and attributes.


{\small
\bibliographystyle{ieee}
\bibliography{egbib}

\begin{thebibliography}{10}\itemsep=-1pt

\bibitem{akata2016multi}
Z.~Akata, M.~Malinowski, M.~Fritz, and B.~Schiele.
\newblock Multi-cue zero-shot learning with strong supervision.
\newblock In {\em {CVPR}}, 2016.

\bibitem{akata2016label}
Z.~Akata, F.~Perronnin, Z.~Harchaoui, and C.~Schmid.
\newblock Label-embedding for image classification.
\newblock {\em PAMI}, 38(7):1425--1438, 2016.

\bibitem{akata2015evaluation}
Z.~Akata, S.~Reed, D.~Walter, H.~Lee, and B.~Schiele.
\newblock Evaluation of output embeddings for fine-grained image
  classification.
\newblock In {\em {CVPR}}, 2015.

\bibitem{IsSM07}
E.~Akturk, G.~Bagci, and R.~Sever.
\newblock Is sharma-mittal entropy really a step beyond tsallis and r{\'e}nyi
  entropies?
\newblock {\em arXiv preprint cond-mat/0703277}, 2007.

\bibitem{changpinyo2016synthesized}
S.~Changpinyo, W.-L. Chao, B.~Gong, and F.~Sha.
\newblock Synthesized classifiers for zero-shot learning.
\newblock In {\em {CVPR}}, pages 5327--5336, 2016.

\bibitem{chao2016empirical}
W.-L. Chao, S.~Changpinyo, B.~Gong, and F.~Sha.
\newblock An empirical study and analysis of generalized zero-shot learning for
  object recognition in the wild.
\newblock In {\em {ECCV}}, pages 52--68. Springer, 2016.

\bibitem{dipaola2009incorporating}
S.~DiPaola and L.~Gabora.
\newblock Incorporating characteristics of human creativity into an
  evolutionary art algorithm.
\newblock {\em Genetic Programming and Evolvable Machines}, 10(2):97--110,
  2009.

\bibitem{dumoulin2016learned}
V.~Dumoulin, J.~Shlens, M.~Kudlur, A.~Behboodi, F.~Lemic, A.~Wolisz,
  M.~Molinaro, C.~Hirche, M.~Hayashi, E.~Bagan, et~al.
\newblock A learned representation for artistic style.
\newblock {\em ICLR}, 2017.

\bibitem{elgammal2017can}
A.~Elgammal, B.~Liu, M.~Elhoseiny, and M.~Mazzone.
\newblock Can: Creative adversarial networks, generating" art" by learning
  about styles and deviating from style norms.
\newblock In {\em International Conference on Computational Creativity}, 2017.

\bibitem{elhoseiny2016write}
M.~Elhoseiny, A.~Elgammal, and B.~Saleh.
\newblock Write a classifier: Predicting visual classifiers from unstructured
  text.
\newblock {\em PAMI}, 2016.

\bibitem{elhoseiny2013write}
M.~Elhoseiny, B.~Saleh, and A.~Elgammal.
\newblock Write a classifier: Zero-shot learning using purely textual
  descriptions.
\newblock In {\em {ICCV}}, 2013.

\bibitem{Elhoseiny_2017_CVPR}
M.~Elhoseiny, Y.~Zhu, H.~Zhang, and A.~Elgammal.
\newblock Link the head to the "beak": Zero shot learning from noisy text
  description at part precision.
\newblock In {\em {CVPR}}, July 2017.

\bibitem{farhadi2009describing}
A.~Farhadi, I.~Endres, D.~Hoiem, and D.~Forsyth.
\newblock Describing objects by their attributes.
\newblock In {\em CVPR 2009.}, pages 1778--1785. IEEE, 2009.

\bibitem{felix2018multi}
R.~Felix, V.~B. Kumar, I.~Reid, and G.~Carneiro.
\newblock Multi-modal cycle-consistent generalized zero-shot learning.
\newblock In {\em ECCV}, pages 21--37, 2018.

\bibitem{frome2013devise}
A.~Frome, G.~S. Corrado, J.~Shlens, S.~Bengio, J.~Dean, T.~Mikolov, et~al.
\newblock Devise: A deep visual-semantic embedding model.
\newblock In {\em {NIPS}}, pages 2121--2129, 2013.

\bibitem{Gatys2016ImageStyleTransfer}
L.~A. Gatys, A.~S. Ecker, and M.~Bethge.
\newblock Image style transfer using convolutional neural networks.
\newblock In {\em CVPR}, 2016.

\bibitem{goodfellow2014generative}
I.~Goodfellow, J.~Pouget-Abadie, M.~Mirza, B.~Xu, D.~Warde-Farley, S.~Ozair,
  A.~Courville, and Y.~Bengio.
\newblock Generative adversarial nets.
\newblock In {\em {NIPS}}, pages 2672--2680, 2014.

\bibitem{gulrajani2017improved}
I.~Gulrajani, F.~Ahmed, M.~Arjovsky, V.~Dumoulin, and A.~Courville.
\newblock Improved training of wasserstein gans.
\newblock {\em arXiv preprint arXiv:1704.00028}, 2017.

\bibitem{guo2017synthesizing}
Y.~Guo, G.~Ding, J.~Han, and Y.~Gao.
\newblock Synthesizing samples for zero-shot learning.
\newblock In {\em {IJCAI}}, 2017.

\bibitem{guo2017zero}
Y.~Guo, G.~Ding, J.~Han, and Y.~Gao.
\newblock Zero-shot learning with transferred samples.
\newblock {\em IEEE Transactions on Image Processing}, 2017.

\bibitem{sm_1976}
H.~Gupta and B.~D. Sharma.
\newblock On non-additive measures of inaccuracy.
\newblock {\em Czechoslovak Mathematical Journal}, 26(4):584--595, 1976.

\bibitem{ha2017neural}
D.~Ha and D.~Eck.
\newblock A neural representation of sketch drawings.
\newblock {\em ICLR}, 2018.

\bibitem{Bhattacharyya}
T.~Kailath.
\newblock The divergence and bhattacharyya distance measures in signal
  selection.
\newblock {\em IEEE transactions on communication technology}, 15(1):52--60,
  1967.

\bibitem{bhatt67}
T.~Kailath.
\newblock The divergence and bhattacharyya distance measures in signal
  selection.
\newblock {\em IEEE transactions on communication technology}, 15(1):52--60,
  1967.

\bibitem{kodirov2017semantic}
E.~Kodirov, T.~Xiang, and S.~Gong.
\newblock Semantic autoencoder for zero-shot learning.
\newblock {\em arXiv preprint arXiv:1704.08345}, 2017.

\bibitem{kumar2018generalized}
V.~Kumar~Verma, G.~Arora, A.~Mishra, and P.~Rai.
\newblock Generalized zero-shot learning via synthesized examples.
\newblock In {\em CVPR}, 2018.

\bibitem{lampert2009}
C.~H. Lampert, H.~Nickisch, and S.~Harmeling.
\newblock Learning to detect unseen object classes by between-class attribute
  transfer.
\newblock In {\em {CVPR}}, pages 951--958. IEEE, 2009.

\bibitem{Lampert2014}
C.~H. Lampert, H.~Nickisch, and S.~Harmeling.
\newblock Attribute-based classification for zero-shot visual object
  categorization.
\newblock {\em PAMI}, 36(3):453--465, March 2014.

\bibitem{lei2015predicting}
J.~Lei~Ba, K.~Swersky, S.~Fidler, et~al.
\newblock Predicting deep zero-shot convolutional neural networks using textual
  descriptions.
\newblock In {\em {ICCV}}, 2015.

\bibitem{long2017zero}
Y.~Long, L.~Liu, L.~Shao, F.~Shen, G.~Ding, and J.~Han.
\newblock From zero-shot learning to conventional supervised classification:
  Unseen visual data synthesis.
\newblock In {\em {CVPR}}, 2017.

\bibitem{long2017describing}
Y.~Long and L.~Shao.
\newblock Describing unseen classes by exemplars: Zero-shot learning using
  grouped simile ensemble.
\newblock In {\em 2017 IEEE Winter Conference on Applications of Computer
  Vision (WACV)}, pages 907--915. IEEE, 2017.

\bibitem{long2017learning}
Y.~Long and L.~Shao.
\newblock Learning to recognise unseen classes by a few similes.
\newblock In {\em Proceedings of the 25th ACM international conference on
  Multimedia}, pages 636--644. ACM, 2017.

\bibitem{machado2000nevar}
P.~Machado and A.~Cardoso.
\newblock Nevar--the assessment of an evolutionary art tool.
\newblock In {\em Proc. of the AISB00 Symposium on Creative \& Cultural Aspects
  and Applications of AI \& Cognitive Science}, volume 456, 2000.

\bibitem{martindale1990clockwork}
C.~Martindale.
\newblock {\em The clockwork muse: The predictability of artistic change.}
\newblock Basic Books, 1990.

\bibitem{mordvintsev2015inceptionism}
A.~Mordvintsev, C.~Olah, and M.~Tyka.
\newblock Inceptionism: Going deeper into neural networks.
\newblock {\em Google Research Blog. Retrieved June}, 2015.

\bibitem{norouzi2013zero}
M.~Norouzi, T.~Mikolov, S.~Bengio, Y.~Singer, J.~Shlens, A.~Frome, G.~S.
  Corrado, and J.~Dean.
\newblock Zero-shot learning by convex combination of semantic embeddings.
\newblock {\em arXiv preprint arXiv:1312.5650}, 2013.

\bibitem{odena2016conditional}
A.~Odena, C.~Olah, and J.~Shlens.
\newblock Conditional image synthesis with auxiliary classifier gans.
\newblock In {\em ICML}, 2017.

\bibitem{patterson2012sun}
G.~Patterson and J.~Hays.
\newblock Sun attribute database: Discovering, annotating, and recognizing
  scene attributes.
\newblock In {\em Computer Vision and Pattern Recognition (CVPR), 2012 IEEE
  Conference on}, pages 2751--2758. IEEE, 2012.

\bibitem{Qiao2016}
R.~Qiao, L.~Liu, C.~Shen, and A.~v.~d. Hengel.
\newblock Less is more: Zero-shot learning from online textual documents with
  noise suppression.
\newblock In {\em {CVPR}}, June 2016.

\bibitem{radford2015unsupervised}
A.~Radford, L.~Metz, and S.~Chintala.
\newblock Unsupervised representation learning with deep convolutional
  generative adversarial networks.
\newblock {\em ICLR}, 2016.

\bibitem{reed2016learning}
S.~Reed, Z.~Akata, B.~Schiele, and H.~Lee.
\newblock Learning deep representations of fine-grained visual descriptions.
\newblock In {\em {CVPR}}, 2016.

\bibitem{Renyi60}
A.~R\'{e}nyi.
\newblock {On Measures Of Entropy And Information}.
\newblock In {\em Berkeley Symposium on Mathematics, Statistics and
  Probability}, 1960.

\bibitem{romera2015embarrassingly}
B.~Romera-Paredes and P.~Torr.
\newblock An embarrassingly simple approach to zero-shot learning.
\newblock In {\em {ICML}}, pages 2152--2161, 2015.

\bibitem{salakhutdinov2011learning}
R.~Salakhutdinov, A.~Torralba, and J.~Tenenbaum.
\newblock Learning to share visual appearance for multiclass object detection.
\newblock In {\em CVPR}, 2011.

\bibitem{salton1988term}
G.~Salton and C.~Buckley.
\newblock Term-weighting approaches in automatic text retrieval.
\newblock {\em Information processing \& management}, 24(5):513--523, 1988.

\bibitem{sbai2018design}
O.~Sbai, M.~Elhoseiny, A.~Bordes, Y.~LeCun, and C.~Couprie.
\newblock Design: Design inspiration from generative networks.
\newblock In {\em ECCV workshop}, 2018.

\bibitem{SM75}
B.~D. Sharma and D.~P. Mittal.
\newblock New nonadditive measures of entropy for discrete probability
  distributions.
\newblock {\em J. Math. Sci}, 10:28--40, 1975.

\bibitem{shigeto2015ridge}
Y.~Shigeto, I.~Suzuki, K.~Hara, M.~Shimbo, and Y.~Matsumoto.
\newblock Ridge regression, hubness, and zero-shot learning.
\newblock In {\em Joint European Conference on Machine Learning and Knowledge
  Discovery in Databases}, pages 135--151. Springer, 2015.

\bibitem{simonyan2014very}
K.~Simonyan and A.~Zisserman.
\newblock Very deep convolutional networks for large-scale image recognition.
\newblock In {\em {ICLR}}, 2015.

\bibitem{socher2013zero}
R.~Socher, M.~Ganjoo, C.~D. Manning, and A.~Ng.
\newblock Zero-shot learning through cross-modal transfer.
\newblock In {\em {NIPS}}, pages 935--943, 2013.

\bibitem{tsai2017learning}
Y.-H.~H. Tsai, L.-K. Huang, and R.~Salakhutdinov.
\newblock Learning robust visual-semantic embeddings.
\newblock In {\em {ICCV}}, 2017.

\bibitem{Tsallis88}
C.~Tsallis.
\newblock {Possible generalization of Boltzmann-Gibbs statistics}.
\newblock {\em J. Statist. Phys.}, 1988.

\bibitem{Horn2015}
G.~Van~Horn, S.~Branson, R.~Farrell, S.~Haber, J.~Barry, P.~Ipeirotis,
  P.~Perona, and S.~Belongie.
\newblock Building a bird recognition app and large scale dataset with citizen
  scientists: The fine print in fine-grained dataset collection.
\newblock In {\em {CVPR}}, 2015.

\bibitem{wah2011caltech}
C.~Wah, S.~Branson, P.~Welinder, P.~Perona, and S.~Belongie.
\newblock The caltech-ucsd birds-200-2011 dataset.
\newblock 2011.

\bibitem{wiki_crested19}
Wikipedia.
\newblock Crested auklet.
\newblock \url{https://en.wikipedia.org/wiki/Crested_auklet}, 2009.
\newblock [Online; accessed 19-March-2019].

\bibitem{xian2016latent}
Y.~Xian, Z.~Akata, G.~Sharma, Q.~Nguyen, M.~Hein, and B.~Schiele.
\newblock Latent embeddings for zero-shot classification.
\newblock In {\em CVPR}, pages 69--77, 2016.

\bibitem{gbu}
Y.~Xian, C.~H. Lampert, B.~Schiele, and Z.~Akata.
\newblock Zero-shot learning-a comprehensive evaluation of the good, the bad
  and the ugly.
\newblock {\em PAMI}, 2018.

\bibitem{xian2018feature}
Y.~Xian, T.~Lorenz, B.~Schiele, and Z.~Akata.
\newblock Feature generating networks for zero-shot learning.
\newblock In {\em CVPR}, 2018.

\bibitem{yang2014unified}
Y.~Yang and T.~M. Hospedales.
\newblock A unified perspective on multi-domain and multi-task learning.
\newblock In {\em {ICLR}}, 2015.

\bibitem{zhang2016spda}
H.~Zhang, T.~Xu, M.~Elhoseiny, X.~Huang, S.~Zhang, A.~Elgammal, and D.~Metaxas.
\newblock Spda-cnn: Unifying semantic part detection and abstraction for
  fine-grained recognition.
\newblock In {\em {CVPR}}, pages 1143--1152, 2016.

\bibitem{zhang2016learning}
L.~Zhang, T.~Xiang, and S.~Gong.
\newblock Learning a deep embedding model for zero-shot learning.
\newblock In {\em {CVPR}}, 2016.

\bibitem{zhang2015zero}
Z.~Zhang and V.~Saligrama.
\newblock Zero-shot learning via semantic similarity embedding.
\newblock In {\em ICCV}, pages 4166--4174, 2015.

\bibitem{Elhoseiny_2018_CVPR}
Y.~Zhu, M.~Elhoseiny, B.~Liu, X.~Peng, and A.~Elgammal.
\newblock A generative adversarial approach for zero-shot learning from noisy
  texts.
\newblock In {\em CVPR}, 2018.

\bibitem{zipf1935human}
G.~K. Zipf.
\newblock The psycho biology of language an introduction to dynamic philology.
\newblock 1935.

\bibitem{zipf1949human}
G.~K. Zipf.
\newblock Human behavior and the principle of least effort.
\newblock 1949.

\end{thebibliography}


\begin{thebibliography}{10}\itemsep=-1pt

\bibitem{akata2016multi}
Z.~Akata, M.~Malinowski, M.~Fritz, and B.~Schiele.
\newblock Multi-cue zero-shot learning with strong supervision.
\newblock In {\em {CVPR}}, 2016.

\bibitem{akata2016label}
Z.~Akata, F.~Perronnin, Z.~Harchaoui, and C.~Schmid.
\newblock Label-embedding for image classification.
\newblock {\em IEEE Transactions on pattern analysis and machine intelligence},
  38(7):1425--1438, 2016.

\bibitem{akata2015evaluation}
Z.~Akata, S.~Reed, D.~Walter, H.~Lee, and B.~Schiele.
\newblock Evaluation of output embeddings for fine-grained image
  classification.
\newblock In {\em {CVPR}}, 2015.

\bibitem{arjovsky2017wasserstein}
M.~Arjovsky, S.~Chintala, and L.~Bottou.
\newblock Wasserstein gan.
\newblock {\em arXiv preprint arXiv:1701.07875}, 2017.

\bibitem{changpinyo2016synthesized}
S.~Changpinyo, W.-L. Chao, B.~Gong, and F.~Sha.
\newblock Synthesized classifiers for zero-shot learning.
\newblock In {\em {CVPR}}, pages 5327--5336, 2016.

\bibitem{chao2016empirical}
W.-L. Chao, S.~Changpinyo, B.~Gong, and F.~Sha.
\newblock An empirical study and analysis of generalized zero-shot learning for
  object recognition in the wild.
\newblock In {\em {ECCV}}, pages 52--68. Springer, 2016.

\bibitem{comaniciu2002mean}
D.~Comaniciu and P.~Meer.
\newblock Mean shift: A robust approach toward feature space analysis.
\newblock {\em IEEE Transactions on pattern analysis and machine intelligence},
  24(5):603--619, 2002.

\bibitem{elhoseiny2016write}
M.~Elhoseiny, A.~Elgammal, and B.~Saleh.
\newblock Write a classifier: Predicting visual classifiers from unstructured
  text.
\newblock {\em IEEE Transactions on pattern analysis and machine intelligence},
  2016.

\bibitem{elhoseiny2013write}
M.~Elhoseiny, B.~Saleh, and A.~Elgammal.
\newblock Write a classifier: Zero-shot learning using purely textual
  descriptions.
\newblock In {\em {ICCV}}, 2013.

\bibitem{Elhoseiny_2017_CVPR}
M.~Elhoseiny, Y.~Zhu, H.~Zhang, and A.~Elgammal.
\newblock Link the head to the "beak": Zero shot learning from noisy text
  description at part precision.
\newblock In {\em {CVPR}}, July 2017.

\bibitem{frome2013devise}
A.~Frome, G.~S. Corrado, J.~Shlens, S.~Bengio, J.~Dean, T.~Mikolov, et~al.
\newblock Devise: A deep visual-semantic embedding model.
\newblock In {\em {NIPS}}, pages 2121--2129, 2013.

\bibitem{fu2016semi}
Y.~Fu and L.~Sigal.
\newblock Semi-supervised vocabulary-informed learning.
\newblock In {\em {CVPR}}, pages 5337--5346, 2016.

\bibitem{girshick2015fast}
R.~Girshick.
\newblock Fast r-cnn.
\newblock In {\em {ICCV}}, pages 1440--1448, 2015.

\bibitem{goodfellow2014generative}
I.~Goodfellow, J.~Pouget-Abadie, M.~Mirza, B.~Xu, D.~Warde-Farley, S.~Ozair,
  A.~Courville, and Y.~Bengio.
\newblock Generative adversarial nets.
\newblock In {\em {NIPS}}, pages 2672--2680, 2014.

\bibitem{gulrajani2017improved}
I.~Gulrajani, F.~Ahmed, M.~Arjovsky, V.~Dumoulin, and A.~Courville.
\newblock Improved training of wasserstein gans.
\newblock {\em arXiv preprint arXiv:1704.00028}, 2017.

\bibitem{guo2017synthesizing}
Y.~Guo, G.~Ding, J.~Han, and Y.~Gao.
\newblock Synthesizing samples for zero-shot learning.
\newblock 2017.

\bibitem{guo2017zero}
Y.~Guo, G.~Ding, J.~Han, and Y.~Gao.
\newblock Zero-shot learning with transferred samples.
\newblock {\em IEEE Transactions on Image Processing}, 2017.

\bibitem{isola2016image}
P.~Isola, J.-Y. Zhu, T.~Zhou, and A.~A. Efros.
\newblock Image-to-image translation with conditional adversarial networks.
\newblock {\em arXiv preprint arXiv:1611.07004}, 2016.

\bibitem{jiang1997semantic}
J.~J. Jiang and D.~W. Conrath.
\newblock Semantic similarity based on corpus statistics and lexical taxonomy.
\newblock {\em arXiv preprint cmp-lg/9709008}, 1997.

\bibitem{lampert2009}
C.~H. Lampert, H.~Nickisch, and S.~Harmeling.
\newblock Learning to detect unseen object classes by between-class attribute
  transfer.
\newblock In {\em Computer Vision and Pattern Recognition, 2009. CVPR 2009.
  IEEE Conference on}, pages 951--958. IEEE, 2009.

\bibitem{Lampert2014}
C.~H. Lampert, H.~Nickisch, and S.~Harmeling.
\newblock Attribute-based classification for zero-shot visual object
  categorization.
\newblock {\em IEEE Transactions on Pattern Analysis and Machine Intelligence},
  36(3):453--465, March 2014.

\bibitem{lei2015predicting}
J.~Lei~Ba, K.~Swersky, S.~Fidler, et~al.
\newblock Predicting deep zero-shot convolutional neural networks using textual
  descriptions.
\newblock In {\em {ICCV}}, 2015.

\bibitem{li2017mmd}
C.-L. Li, W.-C. Chang, Y.~Cheng, Y.~Yang, and B.~P{\'o}czos.
\newblock Mmd gan: Towards deeper understanding of moment matching network.
\newblock {\em arXiv preprint arXiv:1705.08584}, 2017.

\bibitem{long2017zero}
Y.~Long, L.~Liu, L.~Shao, F.~Shen, G.~Ding, and J.~Han.
\newblock From zero-shot learning to conventional supervised classification:
  Unseen visual data synthesis.
\newblock {\em arXiv preprint arXiv:1705.01782}, 2017.

\bibitem{maaten2008visualizing}
L.~v.~d. Maaten and G.~Hinton.
\newblock Visualizing data using t-sne.
\newblock {\em Journal of Machine Learning Research}, 9(Nov):2579--2605, 2008.

\bibitem{mao2016multi}
X.~Mao, Q.~Li, H.~Xie, R.~Y. Lau, and Z.~Wang.
\newblock Multi-class generative adversarial networks with the l2 loss
  function.
\newblock {\em arXiv preprint arXiv:1611.04076}, 2016.

\bibitem{mirza2014conditional}
M.~Mirza and S.~Osindero.
\newblock Conditional generative adversarial nets.
\newblock {\em arXiv preprint arXiv:1411.1784}, 2014.

\bibitem{odena2016conditional}
A.~Odena, C.~Olah, and J.~Shlens.
\newblock Conditional image synthesis with auxiliary classifier gans.
\newblock {\em arXiv preprint arXiv:1610.09585}, 2016.

\bibitem{pathak2016context}
D.~Pathak, P.~Krahenbuhl, J.~Donahue, T.~Darrell, and A.~A. Efros.
\newblock Context encoders: Feature learning by inpainting.
\newblock In {\em {CVPR}}, pages 2536--2544, 2016.

\bibitem{Porter}
M.~F. Porter.
\newblock Readings in information retrieval.
\newblock chapter An Algorithm for Suffix Stripping, pages 313--316. Morgan
  Kaufmann Publishers Inc., San Francisco, CA, USA, 1997.

\bibitem{Qiao2016}
R.~Qiao, L.~Liu, C.~Shen, and A.~v.~d. Hengel.
\newblock Less is more: Zero-shot learning from online textual documents with
  noise suppression.
\newblock In {\em {CVPR}}, June 2016.

\bibitem{reed2016learning}
S.~Reed, Z.~Akata, B.~Schiele, and H.~Lee.
\newblock Learning deep representations of fine-grained visual descriptions.
\newblock In {\em {CVPR}}, 2016.

\bibitem{reed2016generative}
S.~Reed, Z.~Akata, X.~Yan, L.~Logeswaran, B.~Schiele, and H.~Lee.
\newblock Generative adversarial text to image synthesis.
\newblock {\em arXiv preprint arXiv:1605.05396}, 2016.

\bibitem{resnik1995using}
P.~Resnik.
\newblock Using information content to evaluate semantic similarity in a
  taxonomy.
\newblock {\em arXiv preprint cmp-lg/9511007}, 1995.

\bibitem{romera2015embarrassingly}
B.~Romera-Paredes and P.~Torr.
\newblock An embarrassingly simple approach to zero-shot learning.
\newblock In {\em International Conference on Machine Learning}, pages
  2152--2161, 2015.

\bibitem{salimans2016improved}
T.~Salimans, I.~Goodfellow, W.~Zaremba, V.~Cheung, A.~Radford, and X.~Chen.
\newblock Improved techniques for training gans.
\newblock In {\em {NIPS}}, pages 2234--2242, 2016.

\bibitem{salton1988term}
G.~Salton and C.~Buckley.
\newblock Term-weighting approaches in automatic text retrieval.
\newblock {\em Information processing \& management}, 24(5):513--523, 1988.

\bibitem{shigeto2015ridge}
Y.~Shigeto, I.~Suzuki, K.~Hara, M.~Shimbo, and Y.~Matsumoto.
\newblock Ridge regression, hubness, and zero-shot learning.
\newblock In {\em Joint European Conference on Machine Learning and Knowledge
  Discovery in Databases}, pages 135--151. Springer, 2015.

\bibitem{simonyan2014very}
K.~Simonyan and A.~Zisserman.
\newblock Very deep convolutional networks for large-scale image recognition.
\newblock {\em arXiv preprint arXiv:1409.1556}, 2014.

\bibitem{socher2013zero}
R.~Socher, M.~Ganjoo, C.~D. Manning, and A.~Ng.
\newblock Zero-shot learning through cross-modal transfer.
\newblock In {\em Advances in neural information processing systems}, pages
  935--943, 2013.

\bibitem{tsai2017learning}
Y.-H.~H. Tsai, L.-K. Huang, and R.~Salakhutdinov.
\newblock Learning robust visual-semantic embeddings.
\newblock {\em {ICCV}}, 2017.

\bibitem{Horn2015}
G.~Van~Horn, S.~Branson, R.~Farrell, S.~Haber, J.~Barry, P.~Ipeirotis,
  P.~Perona, and S.~Belongie.
\newblock Building a bird recognition app and large scale dataset with citizen
  scientists: The fine print in fine-grained dataset collection.
\newblock In {\em {CVPR}, address = {Boston, MA}, keywords = {}}, 2015.

\bibitem{wah2011caltech}
C.~Wah, S.~Branson, P.~Welinder, P.~Perona, and S.~Belongie.
\newblock The caltech-ucsd birds-200-2011 dataset.
\newblock 2011.

\bibitem{Tao18attngan}
T.~Xu, P.~Zhang, Q.~Huang, H.~Zhang, Z.~Gan, X.~Huang, and X.~He.
\newblock Attngan: Fine-grained text to image generation with attentional
  generative adversarial networks.
\newblock 2018.

\bibitem{yang2014unified}
Y.~Yang and T.~M. Hospedales.
\newblock A unified perspective on multi-domain and multi-task learning.
\newblock {\em arXiv preprint arXiv:1412.7489}, 2014.

\bibitem{He_dehaze_2018}
H.~Zhang and V.~M. Patel.
\newblock Densely connected pyramid dehazing network.
\newblock In {\em {CVPR}}, 2018.

\bibitem{zhang2017image}
H.~Zhang, V.~Sindagi, and V.~M. Patel.
\newblock Image de-raining using a conditional generative adversarial network.
\newblock {\em arXiv preprint arXiv:1701.05957}, 2017.

\bibitem{zhang2016spda}
H.~Zhang, T.~Xu, M.~Elhoseiny, X.~Huang, S.~Zhang, A.~Elgammal, and D.~Metaxas.
\newblock Spda-cnn: Unifying semantic part detection and abstraction for
  fine-grained recognition.
\newblock In {\em {CVPR}}, pages 1143--1152, 2016.

\bibitem{zhang2016stackgan}
H.~Zhang, T.~Xu, H.~Li, S.~Zhang, X.~Wang, X.~Huang, and D.~Metaxas.
\newblock Stackgan: Text to photo-realistic image synthesis with stacked
  generative adversarial networks.
\newblock In {\em {ICCV}}, 2017.

\bibitem{zhang2016learning}
L.~Zhang, T.~Xiang, and S.~Gong.
\newblock Learning a deep embedding model for zero-shot learning.
\newblock {\em {CVPR}}, 2016.

\bibitem{zhang2018txt2img}
Z.~Zhang, Y.~Xie, and L.~Yang.
\newblock Photographic text-to-image synthesis with a hierarchically-nested
  adversarial network.
\newblock In {\em {CVPR}}, 2018.

\bibitem{zhang2018cardiac}
Z.~Zhang, L.~Yang, and Y.~Zheng.
\newblock Translating and segmenting multimodal medical volumes with cycle- and
  shape-consistency generative adversarial network.
\newblock In {\em {CVPR}}, 2018.

\bibitem{zhao2016energy}
J.~Zhao, M.~Mathieu, and Y.~LeCun.
\newblock Energy-based generative adversarial network.
\newblock {\em {ICLR}}, 2017.

\end{thebibliography}
}
\newpage
\begin{appendix}

\section{Parakeet Auklet vs Crested Auklet AUC on CUB dataset (SCS split)}
We hypothesized that our method is better in generalization than standard generative ZSL approaches at L51-151 in the main paper. We conduct an additional experiment to verify this claim by plotting the Seen-Unseen curves for only Parakeet Auklet among the seen classes and Crested Auklet among the unseen classes. {The text description of both Auklets are similar and the key difference is the Crested Auklet is featured with forehead crests,  made of black forward-curving feathers. } We note that the prediction space (T) still includes the 200 CUB species (see Fig~\ref{fig_auklet}), but with a focus on analyzing these two categories.  The AUC for the baseline GAZSL is 0.139 and for our CIZSL (GAZSL + our loss) is 0.27 $\approx$ 100\% relative improvement for discriminating these two classes. This demonstrates how the confusion between those two classes is drastically reduced by using our loss, especially for the unseen Crested Auklet (x-axis). This illustrates the key advantage of our added loss,  doubling the capability of GAZSL from 0.13 AUC to 0.27 AUC  to distinguish between two very similar birds: Parakeet Auklet (Seen class) and Crested Auklet (unseen class), in 200-way classification. 

\begin{figure}[b!]
\vspace{-7mm}
	\centering
   \includegraphics[width=0.5\textwidth,height=6.3cm]{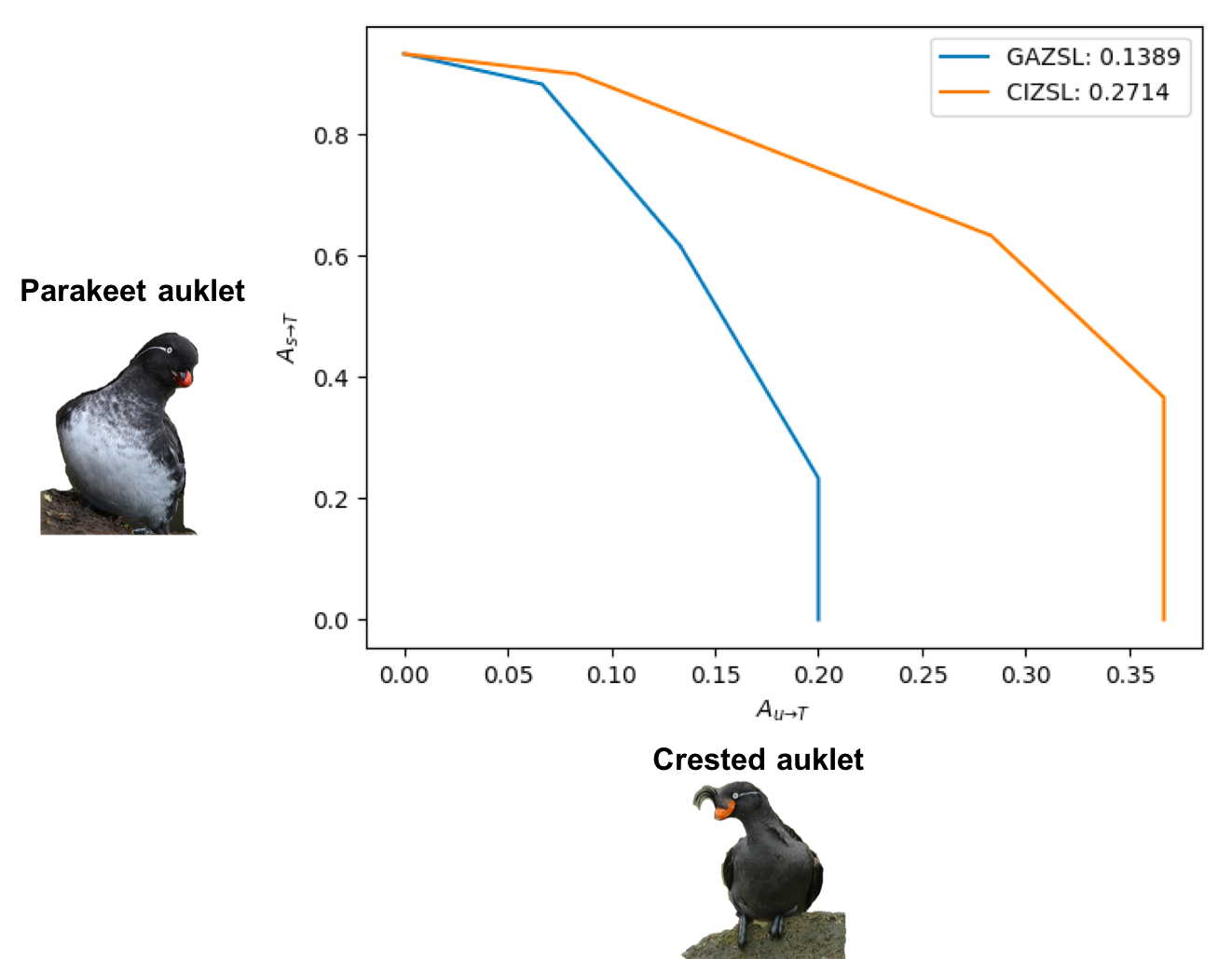}
   \vspace{-2mm}
	\caption{Seen Unseen Curve for Parakeet Auklet (Seen) on the y-axis versus Crested Auklet (unseen) on the x-axis for GAZSL  and CIZSL (GAZSL+our loss) }
	\label{fig_auklet}
\end{figure}
\vspace{-2mm}
\newpage

\section{Divergence Measures}


We generalize the expression of the creativity term to a broader family of divergences, unlocking new way of enforcing deviation from seen classes. 

In \cite{IsSM07}, Sharma-Mittal divergence was studied, originally introduced  \cite{SM75}. Given two parameters ($\alpha$ and $\beta$), 
 the Sharma-Mittal (SM) divergence $SM_{\alpha, \beta}(p\|q)$,  between two distributions $p$ and $q$ is defined $\forall \alpha >0,~\alpha \neq 1,~\beta \neq 1$ as 
\begin{equation}
SM(\alpha, \beta)(p || q) = \frac{1}{\beta-1} \left[ \sum_i (p_i^{1-\alpha} {q_i}^{\alpha})^\frac{1-\beta}{1-\alpha} -1\right]
\label{eq:smgen}
\end{equation}
It was shown in~\cite{IsSM07} that most of the widely used divergence measures are special cases of SM divergence. For instance, each of the R\'enyi, Tsallis and Kullback-Leibler (KL) divergences can be defined as limiting cases of SM divergence as follows:
\begin{equation}
\small
\begin{split}
R_{\alpha}(p\|q) = &\lim_{\beta \to 1} {SM_{\alpha, \beta}(p\|q) }  = \frac{1}{\alpha-1} \ln (\sum_i {p_i^\alpha q_i^{1-\alpha} } )),  \\
T_{\alpha}(p\|q) = &\lim_{\beta \to \alpha} {SM_{\alpha, \beta}(p\|q) }= \frac{1}{\alpha-1} (\sum_i {p_i^\alpha q_i^{1-\alpha} } ) -1),  \\
KL(p\|q) = & \lim_{\beta \to 1, \alpha \to 1} {SM_{\alpha, \beta}(p\|q) } = \sum_i{p_i \ln ( \frac{p_i}{q_i}}).
\end{split}
\label{eqdef}
\end{equation}
In particular, the Bhattacharyya divergence ~\cite{bhatt67}, denoted by $B(p\|q)$   is a limit case of SM and R\'enyi divergences as follows as $\beta \to 1, \alpha \to 0.5$
\begin{equation}
\small
\begin{split}
B(p\|q)& = 2 \lim_{\beta \to 1, \alpha \to 0.5}  SM_{\alpha,\beta}(p\|q)  = - \ln \Big( \sum_i p_i^{0.5} q_i^{0.5} \Big).
\end{split}
\end{equation}
Since the notion of creativity in our work is grounded to maximizing the deviation from existing shapes and textures through KL divergence, we can  generalize our MCE creativity loss by minimizing  Sharma Mittal (SM) divergence   between a uniform distribution and the softmax output $\hat{D}$ as follows
\begin{equation}
\begin{split}
\mathcal{L}_{SM} = SM(\alpha,\beta)( \hat{D} || u) = 
SM(\alpha, \beta)(\hat{D} || u) \\  = \frac{1}{\beta-1} \sum_i (\frac{1}{K}^{1-\alpha} {\hat{D_i}}^{\alpha})^\frac{1-\beta}{1-\alpha} -1
\end{split}
\label{eq_sm}
\end{equation}
\vspace{5mm}

\section{Training Algorithm}

To train our model, we consider visual-semantic feature pairs, images and text, as a joint observation. Visual features are produced either from real data or synthesized by our generator. 
We illustrate in algorithm~\ref{alg_can_label} how $G$ and $D$ are alternatively optimized with an Adam optimizer. The algorithm summarizes the training procedure. In each iteration, the discriminator is optimized for $n_{d}$ steps (lines $6-11$), and the generator is optimized for $1$ step (lines $12-14$). It is important to mention that when $L_e$ has parameters parameters like $\gamma$ and $\beta$ for Sharma-Mittal(SM) divergence, in Eq.~7, that we update these parameters as well by an Adam optimizer and we perform min-max normalization for $L_e$ within each batch to keep the scale of the loss function the same. We denote the parameters of the entropy function as $\theta_E$ (lines $15$). Also, we perform min-max normalization at the batch level for the entropy loss in equation 5

\begin{algorithm}[]
	\begin{algorithmic}[1]
		\STATE{\bfseries Input:} the maximal loops $N_{step}$, the batch size $m$, the iteration number of discriminator in a loop $n_{d}$, the balancing parameter $\lambda_p$, Adam hyperparameters $\alpha_1$, $\beta_1$, $\beta_2$. 
        \FOR{iter $= 1,..., N_{step}$}
        \STATE Sample random text minibatches $t_a, t_b$, noise $z^h$
        \STATE Construct $t^h$ using Eq.6 with different $\alpha$ for each row in the minibatch
        \STATE $\tilde{x}^h \gets G(t^h, z^h)$
		\FOR{$t = 1$, ..., $n_{d}$}
        \STATE Sample a minibatch of images $x$,  matching texts $t$, random noise $z$
		\STATE $\tilde{x}   \gets G(t, z)$
		\STATE Compute the discriminator loss $L_D$ using Eq. 4
        \STATE $\theta_D \gets \text{Adam}(\bigtriangledown_{\theta_D} L_D, \theta_D, \alpha_1, \beta_1, \beta_2)$
        \ENDFOR
        
        \STATE Sample a minibatch of class labels $c$, matching texts $T_c$, random noise $z$
        \STATE Compute the generator loss $L_G$ using Eq. 5

		\STATE\begin{varwidth}[t]{\linewidth}  $\theta_G \gets \text{Adam}(\bigtriangledown_{\theta_G}  L_G ,	\theta, \alpha_1, \beta_1, \beta_2)$ 	
		\end{varwidth}
		\STATE\begin{varwidth}[t]{\linewidth}  $\theta_E \gets \text{Adam}(\bigtriangledown_{\theta_E}  L_G ,	\theta, \alpha_1, \beta_1, \beta_2)$
		\end{varwidth}
		\ENDFOR
	\end{algorithmic}
	\caption{ Training procedure of our approach. We use default values of $n_{d} =5$, $\alpha = 0.001$, $\beta_1 = 0.5$, $\beta_2 =0.9$}
	\label{alg_can_label}
\end{algorithm}

\section{Zero-Shot Retrieval Qualitative Samples}
We show several examples of the retrieval on CUB dataset using SCS split setting. Given a query semantic representation of an unseen class, the task is to retrieve images from this class. Each row is an unseen class. We show three correct retrievals as well as one incorrect retrieval, randomly picked. We note that, even when the method fails to retrieve the correct class, it tends to retrieve visually similar images. For instance, in the Red bellied Woodpecker example (last row in the first subfigure). Our algorithm mistakenly retrieves an image of the red headed woodpecker. It is easy to notice the level of similarity between the two classes, given that both of them are woodpeckers and contain significant red colors on their bodies.

\newpage
\begin{figure*}
    \centering
    \includegraphics[width=15.0cm,height=6.52cm]{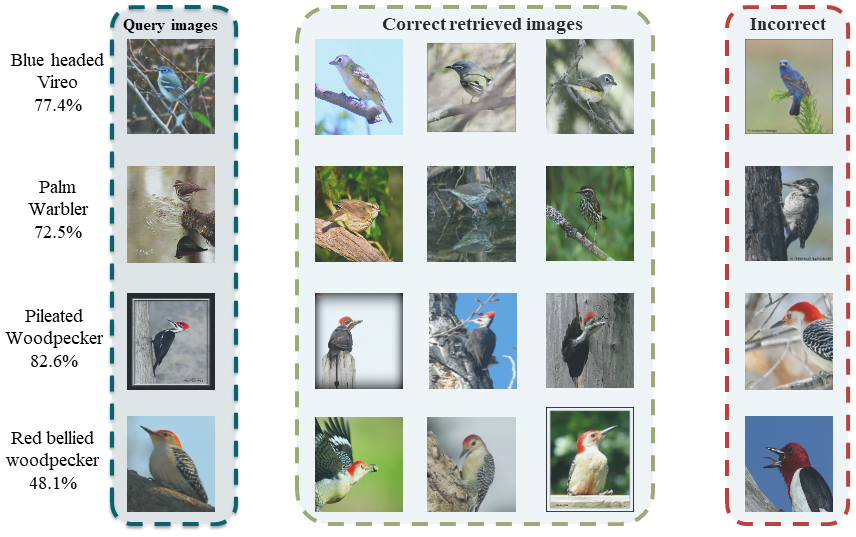}
\end{figure*}
\vspace{-3.27em}
\begin{figure*}
    \centering
    \includegraphics[width=15.0cm,height=6.52cm]{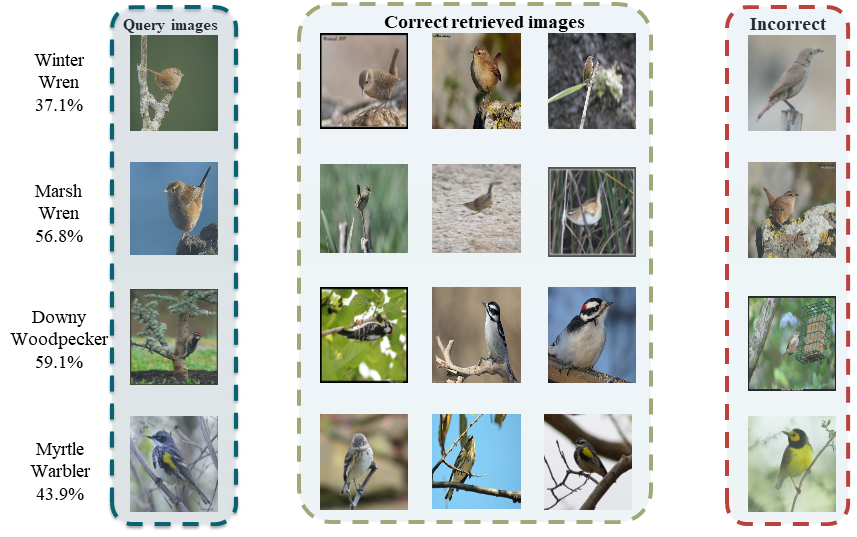}
\end{figure*}
\vspace{-3.27em}
\begin{figure*}
    \centering
    \includegraphics[width=15.0cm,height=6.52cm]{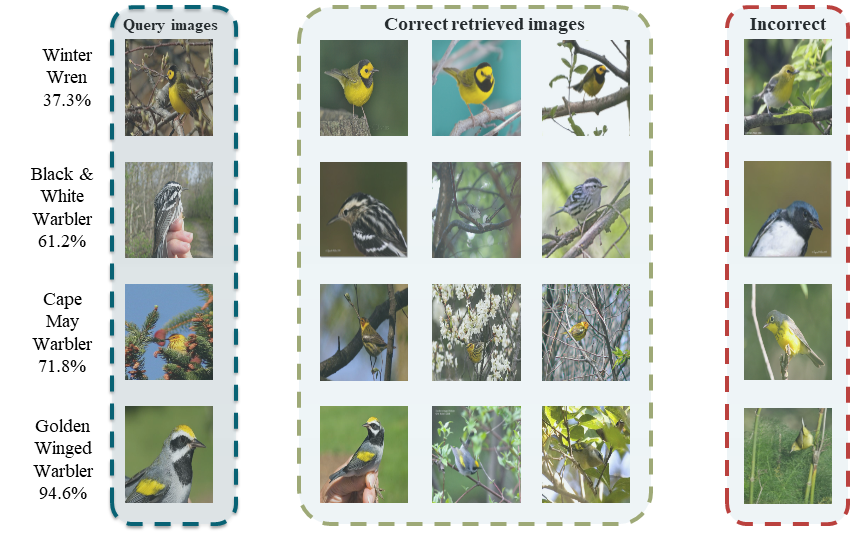}
    \caption{Qualitative results of zero-shot retrieval on CUB dataset using SCS setting.}
\end{figure*}
\newpage

\clearpage
\section{Ablation Study}


In this section we perform an ablation study to investigate best distribution for $\alpha$ in Eq. 6. Unlike our experiments in section 5 of original text where $\lambda$ is cross validated, in this ablation we fix $\lambda$ to examine the effect of changing $\alpha$ distribution on $\alpha$, we achieve better performance.   We observe that when we introduce more variation. Note that generalized Seen-Unseen AUC accuracy is very similar to the results reported in Table 4 of the main paper.

\begin{table}[ht!]
    \centering
        \begin{tabular}{@{}cccccccccc@{}}
        \toprule
        Metric & \multicolumn{4}{c}{Top-1 Accuracy (\%)} &  & \multicolumn{4}{c}{Seen-Unseen AUC (\%)} \\ \cmidrule(lr){2-5} \cmidrule(l){7-10} 
        Dataset & \multicolumn{2}{c}{CUB} & \multicolumn{2}{c}{NAB} &  & \multicolumn{2}{c}{CUB} & \multicolumn{2}{c}{NAB} \\
        Split-Mode & SCS & SCE & SCS & SCE &  & SCS & SCE & SCS & SCE \\ \midrule
        GAZSL~\cite{Elhoseiny_2018_CVPR}- No creative loss & 43.7 & 10.3 & 35.6 & 8.6 &  & 35.4 & 8.7 & 20.4 & 5.8 \\ \midrule
        $\alpha = 0.5$ & \bf 45.7 & \bf 13.9 & 38.6 & 9.1 &  & 39.6 & 11.2 & 24.2 & 6.0 \\
        $\alpha \sim \mathcal{U} (0,1)$ & 45.3 & 13.2 & 38.4 & 9.7 &  & 39.7 & 11.4 & 24.1 & \bf 7.3 \\
        $\alpha \sim \mathcal{U} (0.2,0.8)$ & 45.3 & 13.7 & \bf 38.8 & \bf 9.7 &  & \bf 39.7 & \bf 11.8 & \bf 24.6 & 6.7 \\
        \bottomrule
        \end{tabular}
    \caption{Ablation Study using Zero-Shot recognition on \textbf{CUB} \& \textbf{NAB} datasets with two split settings. We experiment the best $\alpha$ distribution in Eq. 6 of original text.}
    \label{tb:nab_cub_ablation}
\end{table}


\section{Visual Representation}
Zhang \emph{et al.}~\cite{zhang2016spda} showed that fine-grained recognition of bird species can be improved by detecting objects parts and learning a part-based learning representations on top. More specifically, ROI pooling is performed on the  detected bird parts (e.g., wing, head) then semantic features are extracted for each part as a representation.  They named  their network Visual Part Detector/Encoder network (VPDE-net) which has VGG~\cite{simonyan2014very} as backbone architecture. We use the VPDE-net as our feature extractor of images for all our experiments on fine-grained bird recognition data sets, so are all the baselines.

\clearpage
\section{Visualization}
\begin{figure}[b!]
\centering
\includegraphics[width=\textwidth,height=5.8cm]{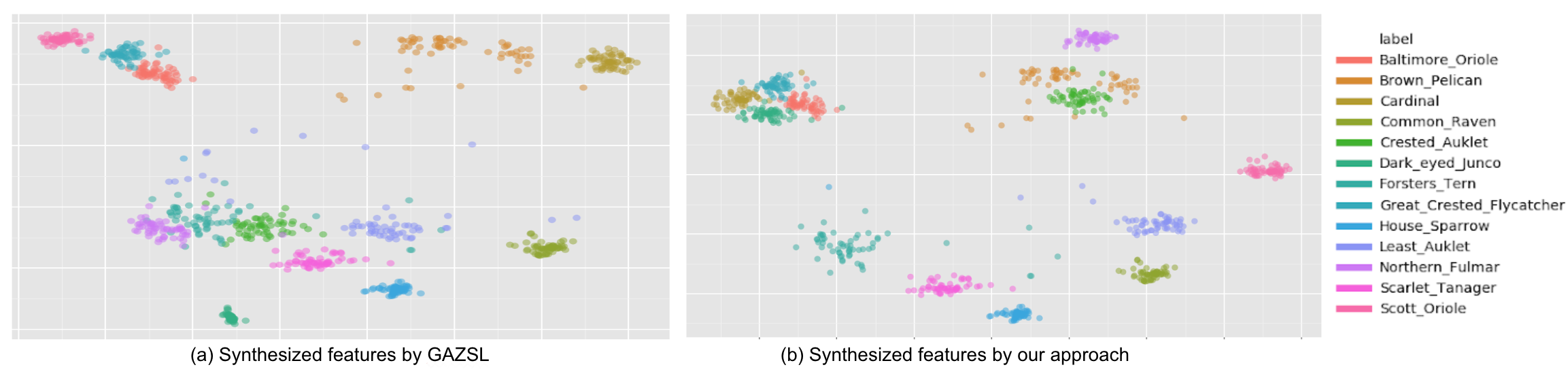}
\caption{t-SNE visualization of features of randomly selected unseen classes. Compared to GAZSL\cite{Elhoseiny_2018_CVPR}, our  method preserves more inter-class discrimination.}
\label{fig:tSNE}
\end{figure}

Our contribution is orthogonal to existing generative zero-shot learning models (e.g., GAZSL~\cite{Elhoseiny_2018_CVPR} and  FeatGen~\cite{xian2018feature}and cycle-(U)WGAN~\cite{felix2018multi}) since it is a learning signal that improves their performance and can be easily integrated to any of them (see  Sec4.2 in the paper). We performed t-SNE visualization of the embeddings for GAZSL with and without our loss as it relates to the learning capability we model; see Fig~\ref{fig:tSNE}.

\end{appendix}
\end{document}